\documentclass[review]{elsarticle}

\usepackage{lineno,hyperref}
\modulolinenumbers[5]
\usepackage[utf8]{inputenc}
\usepackage{graphicx}
\usepackage{float}
\usepackage{pgf}
\usepackage{todonotes}

\journal{Big Data Research}

\def\RR{\textsf{R}\/}

\let\pkg=\strong

\begin{document}
\sloppy

\begin{frontmatter}

\title{Random Forests for Big Data}

\author{Robin Genuer}
\address{INRIA, SISTM team \& ISPED, INSERM U-897, Univ. Bordeaux}
\ead{robin.genuer@isped.u-bordeaux2.fr}

\author{Jean-Michel Poggi}
\address{LMO, Univ. Paris-Sud Orsay \& Univ. Paris Descartes}
\ead{jean-michel.poggi@math.u-psud.fr}

\author{Christine Tuleau-Malot}
\address{Université Côte d'Azur, CNRS, LJAD}
\ead{malot@unice.fr}

\author{Nathalie Villa-Vialaneix}
\address{MIAT, Université de Toulouse, INRA}
\ead{nathalie.villa-vialaneix@inra.fr}

\begin{abstract}
Big Data is one of the major challenges of statistical science and has numerous 
consequences from  algorithmic and theoretical viewpoints. Big Data always 
involve massive data but they also often include online data and data 
heterogeneity. Recently some statistical methods have been adapted to process 
Big Data, like linear regression models, clustering methods and bootstrapping 
schemes. Based on decision trees combined with aggregation and bootstrap ideas, 
random forests were introduced by Breiman in 2001. They are a powerful 
nonparametric statistical method allowing to consider in a single and versatile 
framework regression problems, as well as two-class and multi-class 
classification problems. Focusing on classification problems, this paper proposes
a selective review of available proposals that deal with scaling random forests
to Big Data problems. 
These proposals rely on parallel environments or on online adaptations of 
random forests. We also describe how related quantities -- such as out-of-bag 
error and variable importance -- are addressed in these methods. Then, we 
formulate various remarks for random forests in the Big Data context. Finally, 
we experiment five variants on two massive datasets (15 and 120 millions of 
observations), a simulated one as well as real world data. One variant relies
on subsampling while three others are related to parallel implementations of
random forests and involve either various adaptations of bootstrap to Big Data 
or to ``divide-and-conquer'' approaches. The fifth variant relates on online 
learning of random forests. These numerical experiments lead to highlight the 
relative performance of the different variants, as well as some of their 
limitations.
\end{abstract}

\begin{keyword}
Random Forest\sep Big Data\sep Parallel Computing\sep Bag of Little 
Bootstraps\sep On-line Learning\sep R
\end{keyword}

\end{frontmatter}



\section{Introduction}

\subsection{Statistics in the Big Data world}

{\it Big Data} is one of the major challenges of statistical science and a lot 
of recent references start to think about the numerous consequences of this new 
context from the algorithmic viewpoint and for the theoretical implications of 
this new framework (see \cite{fan_etal_NSR2014,hoerl_etal_WIRCS,jordan_B2013}). 
Big Data always involve massive data but they also often include data streams 
and data heterogeneity (see \cite{besse_etal_p2014} for a general 
introduction), often characterized by the fact that data are frequently not 
structured data, properly indexed in a database and that simple queries cannot 
be easily performed on such data. These features lead to the famous three V 
(Volume, Velocity and Variety) highlighted by the Gartner, Inc., the advisory 
company about information technology research 
\footnote{\href{http://blogs.gartner.com/doug-laney/files/2012/01/%
ad949-3D-Data-Management-Controlling-Data-Volume-Velocity-and-Variety.pdf}{
\url{http://blogs.gartner.com/doug-laney/files/2012/01/}}\\
\href{http://blogs.gartner.com/doug-laney/files/2012/01/%
ad949-3D-Data-Management-Controlling-Data-Volume-Velocity-and-Variety.pdf}{
\url{ad949-3D-Data-Management-Controlling-Data-Volume-Velocity-and-Variety.pdf}}
}. In the most extreme situations, data can even have a too large size to fit 
in a single computer memory. Then data are distributed among several computers. 
For instance, Thusoo {\it et al.} \cite{thusoo_etal_SIGMOD2010} indicate that 
Facebook$^\copyright$ had more than 21PB of data in 2010. Frequently, the 
distribution of such data is managed using specific frameworks dedicated to 
shared computing environments such as Hadoop\footnote{Hadoop, 
\url{http://hadoop.apache.org} is a software environment programmed in Java, 
which contains a file system for distributed architectures (HDFS: Hadoop 
Distributed File System) and dedicated programs for data analysis in parallel 
environments. It has been developed from GoogleFS, The Google File System.}. 

For statistical science, the problem posed by this large amount of data is 
twofold: first, as many statistical procedures have devoted few attention to 
computational runtimes, they can take too long to provide results in an 
acceptable time. When dealing with complex tasks, such as learning of a 
prediction model or complex exploratory analysis, this issue can occur even if  
the dataset would be considered of a moderate size for other (simpler tasks). 
Also, as pointed out in \cite{kane_etal_JSS2013}, the notion of Big Data depends 
itself on the available computing resources. This is especially true when 
relying on the free statistical software \RR{} \cite{RCT_R2016}, massively used 
in the statistical community, which capabilities are strictly limited by RAM. In 
this case, data can be considered as ``large'' if their size exceeds 20\% of RAM 
and as ``massive'' if it exceeds 50\% of RAM, because this amount of data 
strongly limits the available memory for learning the statistical model itself.
As pointed out in \cite{jordan_B2013}, in the near future, statistics will have 
to deal with problems of scale and computational complexity to remain relevant. 
In particular, the collaboration between statisticians and computer scientists 
is needed to control runtimes that will maintain the statistical procedures 
usable on large-scale data while ensuring good statistical properties. 

Recently, some statistical methods have been adapted to process Big Data, 
including linear regression models, clustering methods and bootstrapping schemes 
(see \cite{besse_villavialaneix_p2014} and \cite{wang_etal_p2015} for recent 
reviews and useful references). The main proposed strategies are based on i) 
\emph{subsampling} \cite{badoiu_etal_ACMSTC2002,yan_etal_ICKDDM2009,
kleiner_etal_JRSSB2014,laptev_etal_ICVLDB2012,meng_ICML2013}, ii) \emph{divide 
and conquer approaches} 
\cite{chu_etal_NIPS2010,chen_xie_SS2014,delrio_etal_IS2014}, which consist in 
splitting the problem into several smaller problems and in gathering the 
different results in a final step, iii) \emph{algorithm weakening} 
\cite{chandrasekaran_jordan_PNASUSA2013}, which explicitly treats the trade-off 
between computational time and statistical accuracy using a hierarchy of 
methods with increasing complexity, iv) \emph{online} updates 
\cite{saffari_etal_ICCV2009,denil_etal_ICML2013}, which update the results with 
sequential steps, each having a low computational cost. However, only a few 
papers really address the question of the difference between the ``small data''
standard framework compared to the Big Data in terms of statistical accuracy. 
Noticeable exceptions are the article of Kleiner {\it et al.} 
\cite{kleiner_etal_JRSSB2014} who prove that their ``Bag of Little Bootstraps''
method is statistically equivalent to the standard bootstrap, the article of 
Chen and Xie \cite{chen_xie_SS2014} who demonstrate asymptotic equivalence to 
their ``divide-and-conquer'' based estimator with the estimator based on all 
data in the setting of regression and the article of Yan {\it et al.} 
\cite{yan_etal_ICKDDM2009} who show that the mis-clustering rate of their 
subsampling approach, compared to what would have been obtained with a direct 
approach on the whole dataset, converges to zero when the subsample size grows 
(in an unsupervised setting).

\subsection{Random forests and Big Data}

Based on decision trees and combined with aggregation and bootstrap ideas,
random forests (abbreviated RF in the sequel), were introduced by Breiman
\cite{breiman_ML2001}. They are a powerful nonparametric statistical method 
allowing to consider regression problems as well as two-class and multi-class 
classification problems, in a single and versatile framework. The consistency 
of RF has recently been proved by Scornet {\it et al.} 
\cite{scornet_etal_AS2015}, to cite the most recent result. On a practical 
point of view, RF are widely used (see 
\cite{verikas_etal_PR2011,ziegler_konig_WIRDMKD2014} for recent surveys) and 
exhibit extremely high performance with only a few parameters to tune. 
Since RF are based on the definition of several independent trees, it is thus 
straightforward to obtain a parallel and faster implementation of the RF 
method, in which many trees are built in parallel on different cores.
In addition to the parallel construction of a lot of models
(the trees of a given forest) RF include intensive resampling and, 
it is natural to think about using parallel processing and to consider adapted 
bootstrapping schemes for massive online context. 

Even if the method has already been adapted and implemented to handle Big Data 
in various distributed environments (see, for instance, the libraries 
Mahout \footnote{\url{https://mahout.apache.org}} or MLib, the latter for the 
distributed framework ``Spark''\footnote{\url{https://spark.apache.org/mllib}}, 
among others), a lot of questions remain open.
In this paper, we do not seek to make an exhaustive description of 
the various implementations of RF in scalable environments but we will 
highlight 
some problems posed by the Big Data framework, describe several standard 
strategies that can be use for RF and describe their main features, drawbacks 
and differences with the original approach. We finally experiment five variants 
on two massive datasets (15 and 120 millions of observations), a simulated one 
as well as real world data. One variant relies on subsampling while three 
others are related to parallel implementations of random forests and involve 
either various adaptations of bootstrap to Big Data or to 
``divide-and-conquer'' approaches. The fifth variant relates to 
online learning of RF. 

Since the free statistical software \RR{} \cite{RCT_R2016}, is \textit{de 
facto} the esperanto in the statistical community, and since the most widely 
used programs for designing random forests are also available in \RR{}, we have 
adopted it for numerical experiments as much as possible. More precisely, the 
\RR{} package \pkg{randomForest}, implementing the original RF algorithm using 
Breiman and Cutler's Fortran code, contains many options together with a 
detailed documentation. It has then been used in almost all experiments. The 
only exception is for online RF for which no implementation in \RR{} is 
available. We then use a python library, as an alternative tool in order to 
provide the means to compare this approach to the alternative Big Data variants.

The paper is organized as follows. After this introduction, we briefly recall 
some basic facts about RF in Section~\ref{rf}. 
Then, Section~\ref{scaling-rf-bd} 
is focused on strategies for scaling random forests to Big Data: some proposals 
about RF in parallel environments are reviewed, as well as a description of 
online strategies. The section includes a comparison of the features of every 
method and a discussion about the estimation of the out-of-bag error in these 
methods. 
Section~\ref{experiments} is devoted to numerical experiments on two 
massive datasets, an extensive study on a simulated one and an application to a 
real world one. Finally, Section \ref{conclusion} collects some conclusions and 
discusses two open perspectives.

\section{Random Forests}
\label{rf}

\begin{figure}
	\includegraphics[width=\linewidth]{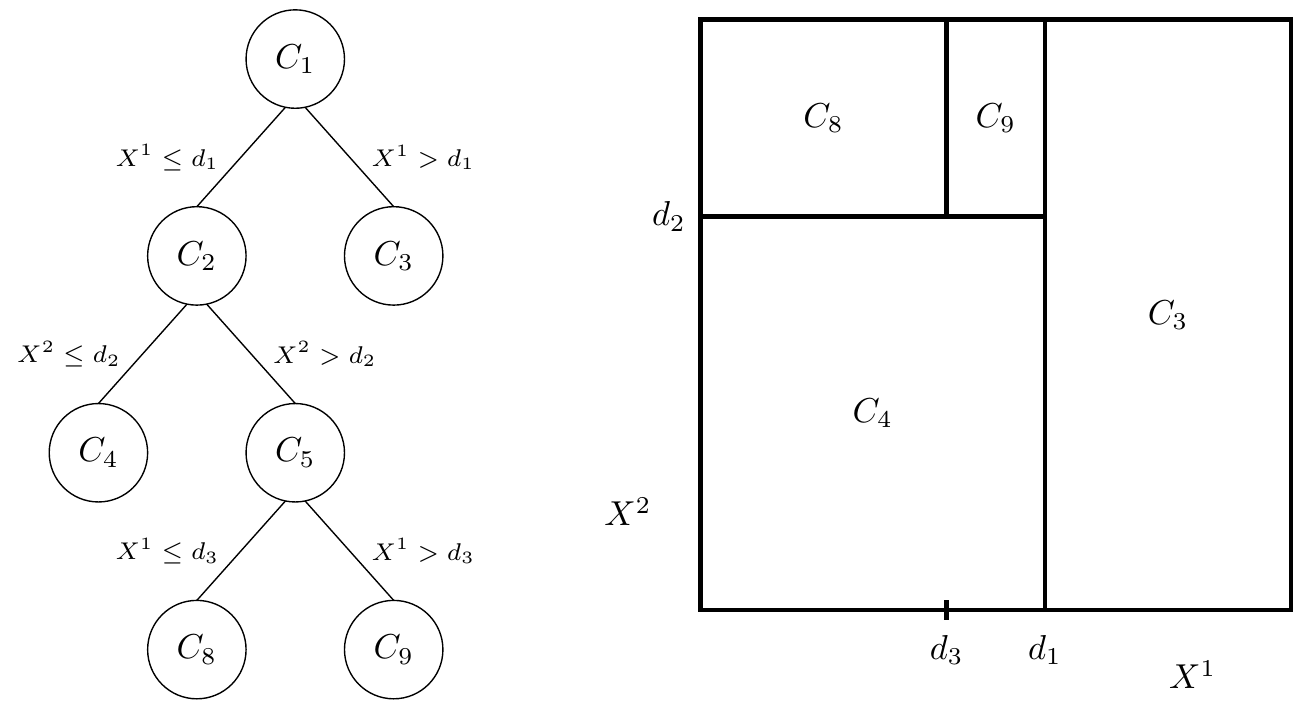}
	\caption{Left: a classification tree allowing to predict the class label
corresponding to a given $x$-value. Right: the associated partition of the 
predictor space.}
	\label{fig::tree}
\end{figure}

Denoting by $L=\{(x_1,y_1),\ldots,(x_n,y_n)\}$ a learning set of independent 
observations of the random vector $(X,Y)$, we distinguish $X=(X^1,...,X^p)$ 
where $X\in{R}^p$ is the vector of the predictors (or explanatory variables) 
from  $Y\in\mathcal{Y}$ the explained variable, where $Y$ is either a class 
label for classification problems or a numerical response for regression ones.
A classifier $s$ is a mapping $s:{R}^p\rightarrow \mathcal{Y}$ while the 
regression function appears naturally to be the function $s$ when we suppose 
that $Y = s(X) + \varepsilon$ with $E[\varepsilon|X]=0$. RF provide estimators 
of either the Bayes classifier, which minimizes the classification error $P(Y 
\neq s(X))$ or the regression function (see 
\cite{bishop_PRML2006,hastie_etal_ESL2009} for further details on 
classification 
and regression problems). RF are a learning method for classification and 
regression based on the CART (Classification and Regression Trees) method 
defined by Breiman {\it et al.} \cite{breiman_etal_CRT1984}. The left part of 
Figure~\ref{fig::tree} provides an example of classification tree. Such a tree 
allows to predict the class label corresponding to a given $x$-value by simply 
starting from the root of the tree (at the top of the left part of the figure) 
and by answering the questions until a leaf is reached. The predicted class is 
then the value labeling the leaf. Such a tree is a classifier $s$ which allows 
to predict a $y$-value for any given $x$-value. This classifier is the function 
which is piecewise constant on the partition described in the right part of 
Figure~\ref{fig::tree}. Note that splits are parallel to the axes defined by 
the original variables leading to an additive model.

\begin{figure}
	\includegraphics[width=\linewidth]{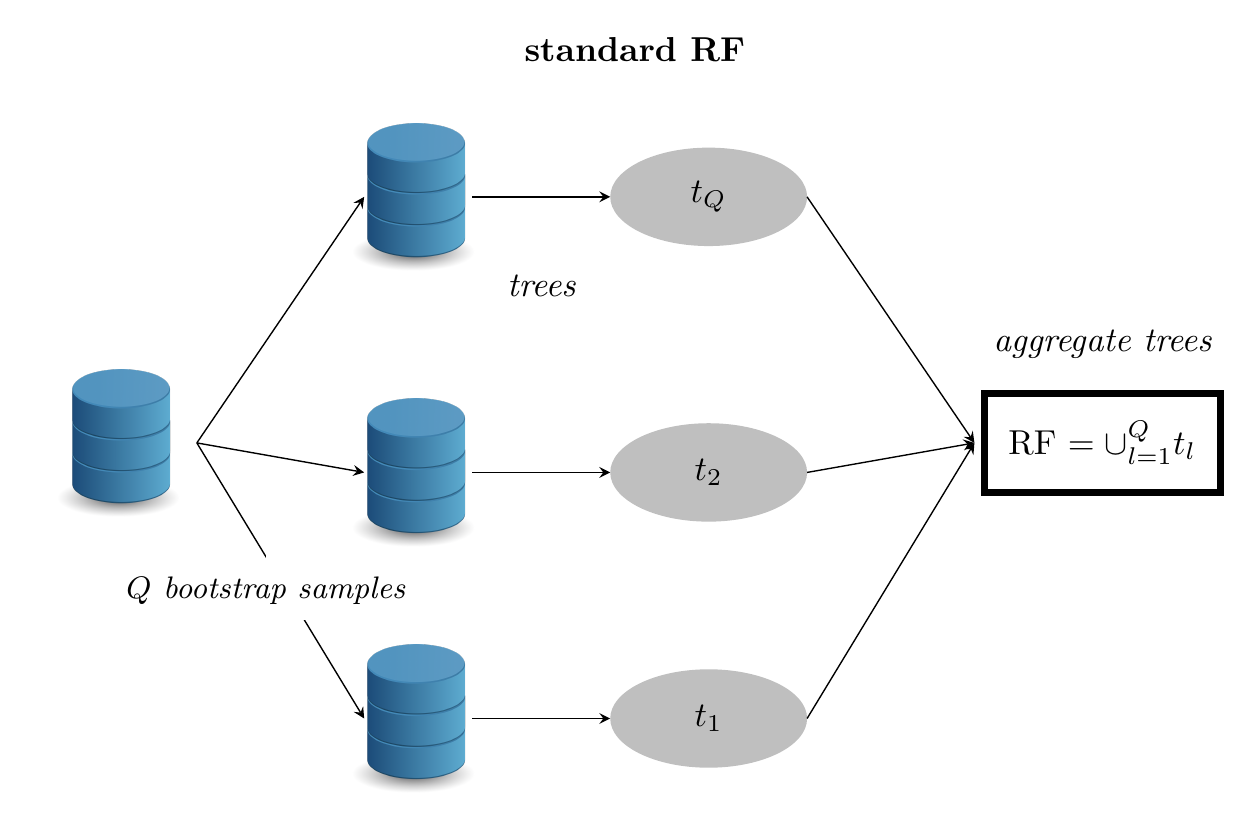}
	\caption{RF construction scheme: starting from the dataset (left of the 
figure), generate bootstrap samples (by randomly selecting $n$ observations
with replacement from learning set $L$) and learn corresponding randomized
binary  decision trees. Finally aggregate them.} 
	\label{fig::rf-scheme}
\end{figure}

While CART is a well-known way to design optimal single trees by performing 
first a growing step and then a pruning one, the principle of RF is to 
aggregate many binary decision trees coming from two random perturbation 
mechanisms: the use of bootstrap samples (obtained by randomly selecting
$n$ observations with replacement from learning set $L$) instead of the whole
sample $L$
and the construction of a randomized tree predictor instead of CART on each 
bootstrap sample. For regression problems, the aggregation step consists in
averaging individual tree predictions, while for classification problems, it
consists in performing a majority vote among individual tree predictions. The 
construction is summarized in Figure~\ref{fig::rf-scheme}.

However, trees in RF have two main differences with respect to CART trees: 
first, in the growing step, at each node, a fixed number of input variables are 
randomly chosen and the best split is calculated only among them, and secondly, 
no pruning is performed.

In the next section, we will explain that most proposals made to adapt RF to 
Big Data often consider the original RF proposed by Breiman as an object that 
simply has to be mimicked in the Big Data context. But we will see, later in 
this article, that alternatives to this vision are possible. Some of these 
alternatives rely on other ways to resample the data and others are based on 
variants in the construction of the trees.

We will concentrate on the prediction performance of RF, focusing on out-of-bag 
(OOB) error, which allows to quantify the variable importance (VI in the 
sequel). The quantification of the variable importance is crucial for many 
procedures involving RF, {\it e.g.}, for ranking the variables before a 
stepwise variable selection strategy (see \cite{genuer_etal_PRL2010}). 
Notations used in this section are given in Table~\ref{table::notations_vi}

\begin{table}[H]
	\centering
	{\small \begin{tabular}{l|c}
		notation & used for\\
		\hline
		$n$ & number of observations in dataset\\
		$Q$ & number of trees in the RF classifier\\
		$t$ & a tree in the RF classifier\\
		OOB$_t$ & set of observations out-of-bag for the tree $t$\\
		errTree$_t$ & misclassification rate for observations in OOB$_t$ made by 
$t$\\
		$\widetilde{\textrm{errTree}}_t^j$ & misclassification rate for 
observations OOB for $t$ \\
		& after a random permutations of values of $X^j$\\
		$\hat{y}_i$ & OOB prediction of observation $x_i$\\
		& (aggregation of predictions made by trees $t$ such that $i \in 
\textrm{OOB}_t$)\\
		errForest & OOB misclassification rate for the RF classifier\\
		VI($X^j$) & Variable importance of $X^j$
	\end{tabular}}
	\caption{Notations used in Section~\ref{rf}.}
	\label{table::notations_vi}
\end{table}

For each tree $t$ of the forest, consider the associated $\mbox{OOB}_t$ sample 
(composed of data not included in the bootstrap sample used to construct $t$).
The OOB error rate of the forest is defined, in the classification case, by:
\begin{equation}
\label{def-errforest}
\mbox{errForest}=\frac{1}{n} \mathrm{Card}\left\{ i \in \{1,\ldots,n\} \: | \: 
y_i \neq \hat{y_i} \right\}
\end{equation}
where $\hat{y_i}$ is the most frequent label predicted by trees $t$ for which 
observation $i$ is in the associated $\mbox{OOB}_t$ sample.

Denote by $\mbox{errTree}_t$ the error (misclassification rate for 
classification) of tree $t$ on its associated $\mbox{OOB}_t$ sample. Now, 
randomly permute the values of $X^j$ in $\mbox{OOB}_t$ to get a perturbed 
sample and compute $\widetilde{\mbox{errTree}_t}^j$, the error of tree $t$ on 
the perturbed sample. Variable importance of $X^j$ is then equal to:
\[
\mbox{VI}(X^j) = \frac{1}{Q} \sum_{t} 
(\widetilde{\mbox{errTree}_t}^j - \mbox{errTree}_t)\
\]
where the sum is over all trees $t$ of the RF and $Q$ denotes the
number of trees of the RF.

\section{Scaling random forests to Big Data}
\label{scaling-rf-bd}

This section discusses the different strategies that can be used to scale 
random forest to Big Data: the first one is subsampling, denoted by {\bf sampRF} in the sequel. Then, 
four parallel implementations of random forests ({\bf parRF}, {\bf moonRF}, {\bf 
blbRF} and {\bf dacRF}), relying on standard parallelization, adaptation of 
bootstraping schemes to Big Data or on a divide-and-conquer approach, are also 
presented. Finally, a different (and not equivalent) approach based on the 
online processing of data is also described, {\bf onRF}. All these variants are 
compared to the original method,  {\bf seqRF}, in which 
all bootstrap samples and trees are built sequentially. The names of the 
different methods and references to the sections in which they are discussed 
are summarized in Table~\ref{table::method-names}.
\begin{table}[H]
	\centering
	{\small \begin{tabular}{l|c|c|c}
		short name & full name & described in & relies on\\
		\hline
		{\bf seqRF} & sequential RF & \ref{rf} & original method\\
		{\bf sampRF} & sampling RF & \ref{sampling} & subsampling\\
		{\bf parRF} & parallel RF & \ref{parallel-rf} & parallelization\\
		{\bf moonRF} & $m$-out-of-$n$ RF & \ref{bootstrap-rf} & Big Data bootstrap\\
		{\bf blbRF} & Bag of Little Bootstraps RF & \ref{bootstrap-rf} & Big Data 
bootstrap\\
		{\bf dacRF} & divide-and-conquer RF & \ref{dac-rf} & divide-and-conquer \\
		{\bf onRF} & online RF & \ref{online-rf} & online learning
	\end{tabular}}
	\caption{Names and references of the different variants of RF described in 
this article.}
	\label{table::method-names}
\end{table}

In addition, the section will use the following notations: RF will denote the 
random forest method (in a generic sense) or the final random forest classifier 
itself, obtained from the various approaches described in this section. The 
number of trees in the final classifier RF is denoted by $Q$, $n$ is the number 
of observations of the original dataset and, when a subsample is taken in this 
dataset (either with or without replacement), it is denoted by $\tau_l$ ($l$ 
identifies the subsample when several subsamples are used) and its size is 
usually denoted by $m$. When different processes are run in parallel, the 
number of processes is denoted by $K$. Depending on the method, this can lead to 
learn smaller RF with $q < Q$ trees that are denoted by $\textrm{RF}_l^{(q)}$, 
in which $l$ is an index that identifies the RF. The notation $\cup_{l=1}^K 
\textrm{RF}_l^{(q)}$ will be used for the classifier obtained from the 
aggregation of $K$ RF with $q$ trees each into a RF with $qK$ trees. Similarly, 
$t_l$ or $t_{ll'}$ denote a tree, identified by the index $l$ or by two 
indices, $l$ and $l'$, when required, and $\cup_{l=1}^q t_l$ denotes the random 
forest obtained from the aggregation of the $q$ trees $t_1$, \ldots, $t_q$. 
Additional notations used in this section are summarized in 
Table~\ref{table::notations}.
\begin{table}[H]
	\centering
	{\small \begin{tabular}{l|c}
		notation & used for\\
		\hline
		$\tau_l$ & subsample of the observations in the dataset\\
		$m$ & number of observations in subsamples\\
		RF & final random forest classifier\\
		$Q$ & number of trees in the final random forest classifier\\
		$K$ & number of processes run in parallel\\
		$q$ & number of trees in intermediate (smaller) random forests\\
		$\textrm{RF}_l^{(q)}$ & RF number $l$ with $q$ trees\\
		$\cup_{l=1}^K \textrm{RF}_l^{(q)}$ & aggregation of $K$ RF with $q$ trees 
in a single classifier\\
		$t_l$ or $t_{ll'}$ & tree identified by the index $l$ or by indices $l$ and 
$l'$\\
		$\cup_{l=1}^q t_l$ & aggregation of $q$ trees in an RF classifier
	\end{tabular}}
	\caption{Notations used in Section~\ref{scaling-rf-bd}.}
	\label{table::notations}
\end{table}

\subsection{Sub-sampling RF (sampRF)}
\label{sampling}

\begin{figure}[H]
	\centering
  \includegraphics[width=0.8\linewidth]{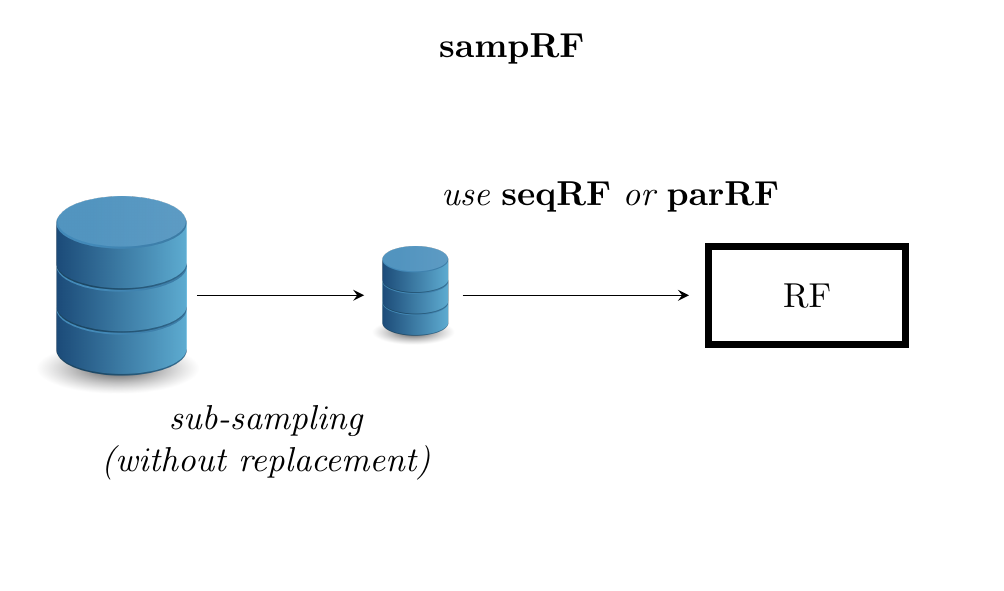}
	\caption{Sub-sampling RF ({\bf sampRF}): $m$ observations out of $n$ are 
randomly selected without replacement and the original RF algorithm ({\bf 
seqRF}) or its parallel version ({\bf parRF}) described in 
Section~\ref{parallel-rf} are used to obtain a final random forest with $Q$ 
trees.}
	\label{fig::sampRF}
\end{figure}

Meng \cite{meng_ICML2013} points the fact that using all data is probably not 
required to obtain accurate estimations in learning methods and that sampling 
approaches is an important approach to deal with Big Data. The natural idea 
behind sampling is to simply subsample $m$ observations out of $n$ without 
replacement in the original sample (with $m \ll n$) and to use the original 
algorithm (either {\bf seqRF} or the parallel implementation, {\bf parRF}, 
described in Section~\ref{parallel-rf}) to process this subsample. This 
method is illustrated in Figure~\ref{fig::sampRF}.

Subsampling is a natural method for statisticians and it is appealing since it 
strongly reduces memory usage and computational efforts. However, it can lead 
to serious biases if the subsample is not carefully designed. More 
precisely, the need to control the representativeness of the subsampling is 
crucial. Random subsampling are usually adequate for such tasks, providing the 
fact that the sampling fraction is large enough. However, in the Big Data 
world, datasets are frequently not structured and indexed. In this situation, 
random subsampling can be a difficult task (see \cite{meng_ICML2013} for a 
discussion on this point and a description of a parallel strategy to overcome 
this problem). Section~\ref{experiments} provides various insights on the 
efficiency of subsampling, on the effect of the sampling fraction and on the 
representativeness of the subsample on the accuracy of the obtained classifier. 
The next section investigates approaches which try to make use of a wider 
proportion of observations in the dataset using efficient computational 
strategies.

\subsection{Parallel implementations of random forests}
\label{parallel-rf}

\begin{figure}[H]
  \includegraphics[width=\linewidth]{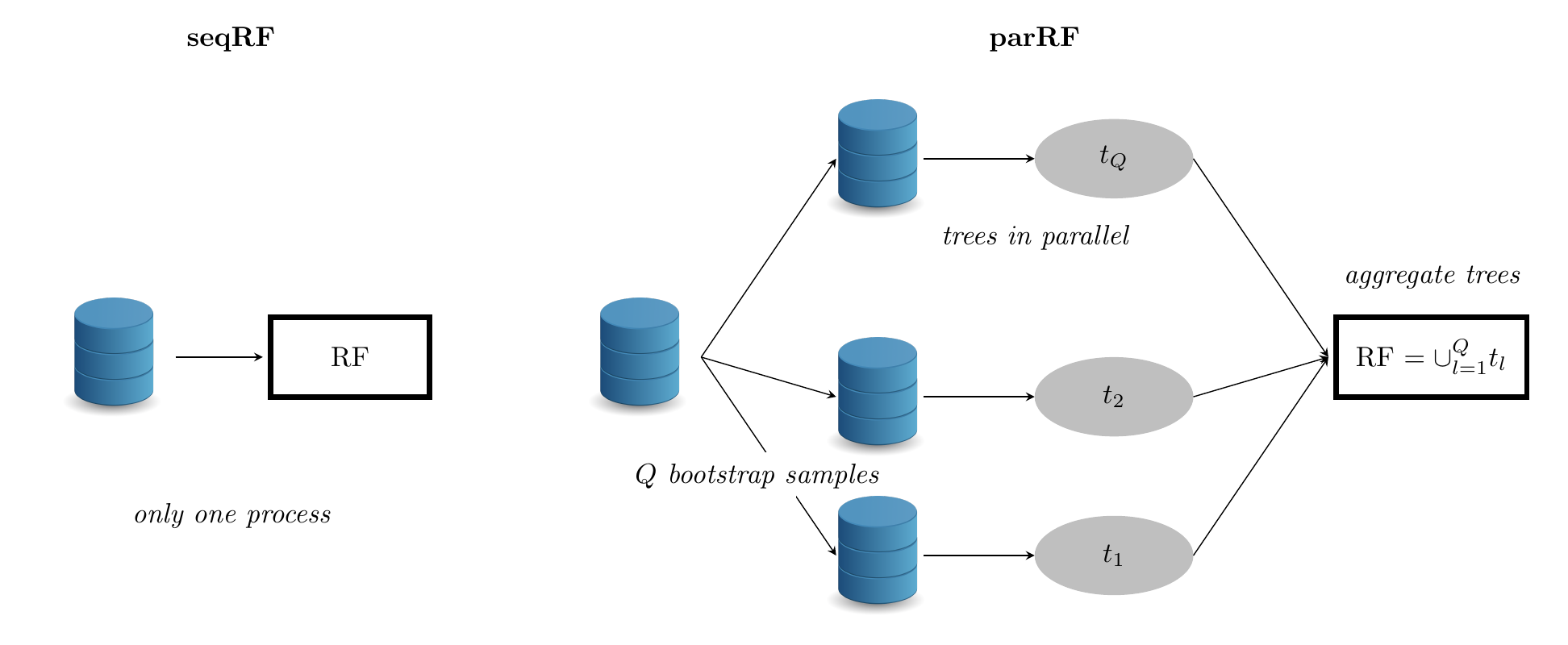}
	\caption{Sequential (left) and parallel (right) implementations of the 
standard RF algorithm. RF is the final random forest with $Q$ trees.
{\bf parRF} builds $K$ small random forests, $\textrm{RF}^{(q)}_l$, with
$q = \frac{Q}{K}$ trees each, using $K$ processes run in parallel.}
	\label{fig::seqANDparRF}
\end{figure}

As pointed in the introduction, RF offer a natural framework for handling Big 
Data. Since the method relies on bootstraping and independant construction of 
many trees, it is naturally suited for parallel computation. Instead of 
building all $Q$ bootstrap samples and trees sequentially as in {\bf seqRF},
bootstrap samples and trees (or sets of a small number of bootstrap samples and 
trees) can be built in parallel. In the sequel, we will denote by {\bf parRF} 
the approach in which $K$ processes corresponding to the learning of a forest 
with $q=\frac{Q}{K}$ trees each are processed in parallel. {\bf seqRF} and
{\bf parRF} implementations are illustrated in Figure~\ref{fig::seqANDparRF} 
(left and right, respectively). Using the {\bf parRF} approach, one can hope for 
a computational time factor decrease of approximately $K$ between {\bf seqRF} 
and {\bf parRF}.

However, as pointed in \cite{kleiner_etal_JRSSB2014}, since the expected size of
a bootstrap sample built from $\{1, \ldots, n\}$ is approximately $0.63n$, the 
need to process hundreds of such samples is hardly feasible in practice when 
$n$ is very large. Moreover, in the original algorithm from 
\cite{breiman_ML2001}, the trees that composed the forest are fully developed 
trees, which means that the trees are grown until every terminal node (leaf) is 
perfectly homogeneous regarding the values of $Y$ for the observations that fall 
in this node. When $n$ is large, and especially in the regression case, this 
leads to very deep trees which are all computationally very expensive and even 
difficult to use for prediction purpose. However, as far as we know, no study 
addresses the question of the impact of controlling and/or tuning the maximum 
number of nodes in the forest's trees.

The next subsection presents alternative solutions to address the issue of 
large size bootstrap samples while relying on the natural parallel background 
of RF. More precisely, we will discuss alternative bootstrap schemes for RF 
($m$-out-of-$n$ bootstrap RF, {\bf moonRF}, and Bag of Little Bootstraps RF, 
{\bf blbRF}) and divide-and-conquer approach, {\bf dacRF}. A last subsection 
will describe and comment on the mismatches of each of these approaches with the 
standard RF method, {\bf seqRF} or {\bf parRF}.

\subsubsection{Alternative bootstrap schemes for RF (moonRF and blbRF)}
\label{bootstrap-rf}

\begin{figure}[H]
  \includegraphics[width=\linewidth]{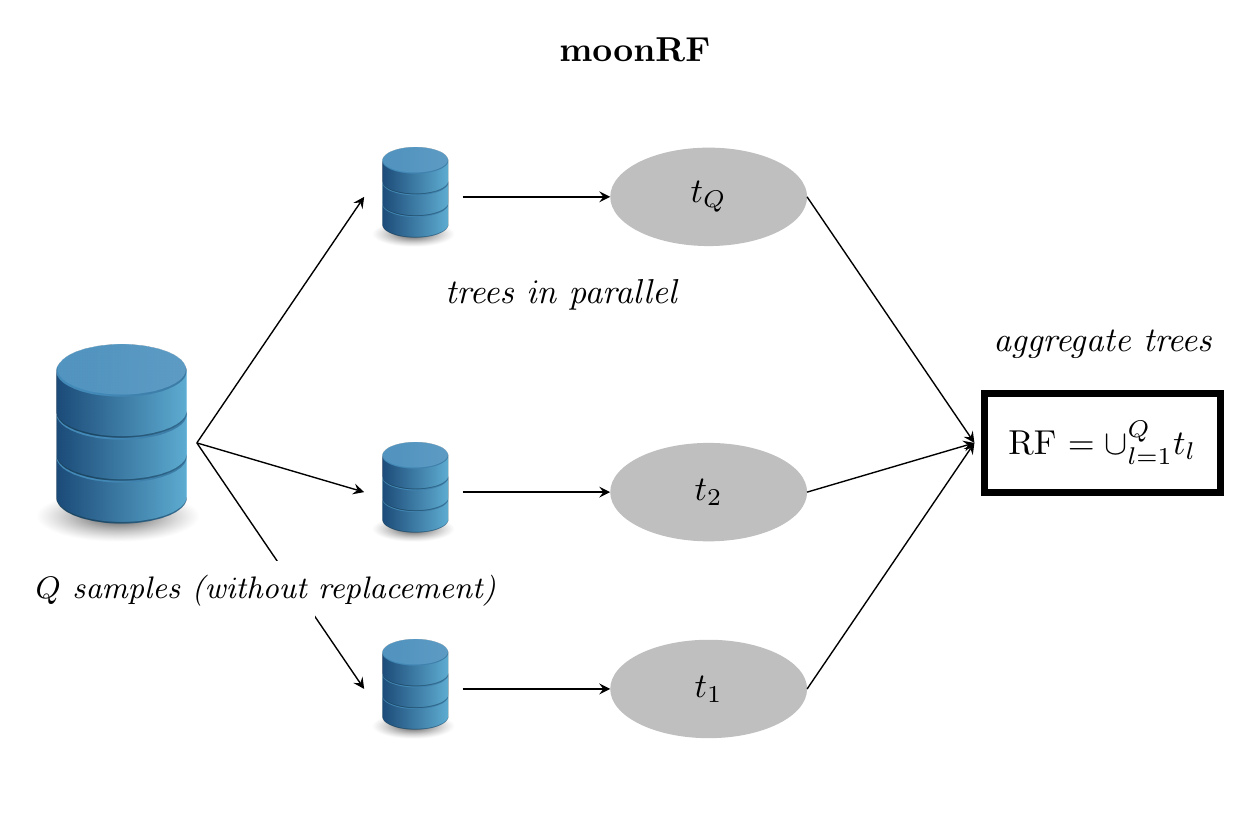}
	\caption{$m$-out-of-$n$ RF ({\bf moonRF}): $Q$ samples without replacement
	with $m$ observations out of $n$ are randomly built in parallel and a tree is 
learned 
from each of these samples. The $Q$ trees are then aggregated to obtain a final 
random forest with $Q$ trees.}
	\label{fig::moonRF}
\end{figure}

To avoid selecting only some of the observations in the original big dataset 
as it is done in {\bf sampRF} (Figure~\ref{fig::sampRF}), 
some 
authors have focused on alternative bootstrap schemes aiming at reducing the 
number of different observations of each bootstrap samples. 
\cite{bickel_etal_SS1997} propose the $m$-out-of-$n$ bootstrap that consists in 
building bootstrap samples with only $m$ observations taken without replacement 
in $\{1, \ldots, n\}$ (for $m \ll n$). This method is illustrated in 
Figure~\ref{fig::moonRF}.
Initially designed to address the computational burden of standard 
bootstrapping, the method performance is strongly dependent on a convenient 
choice of $m$ and the data-driven scheme proposed in \cite{bickel_sakov_SS2008} 
for the selection of $m$ requires to test several different values of $m$ and 
eliminates computational gains.

\begin{figure}[H]
  \includegraphics[width=\linewidth]{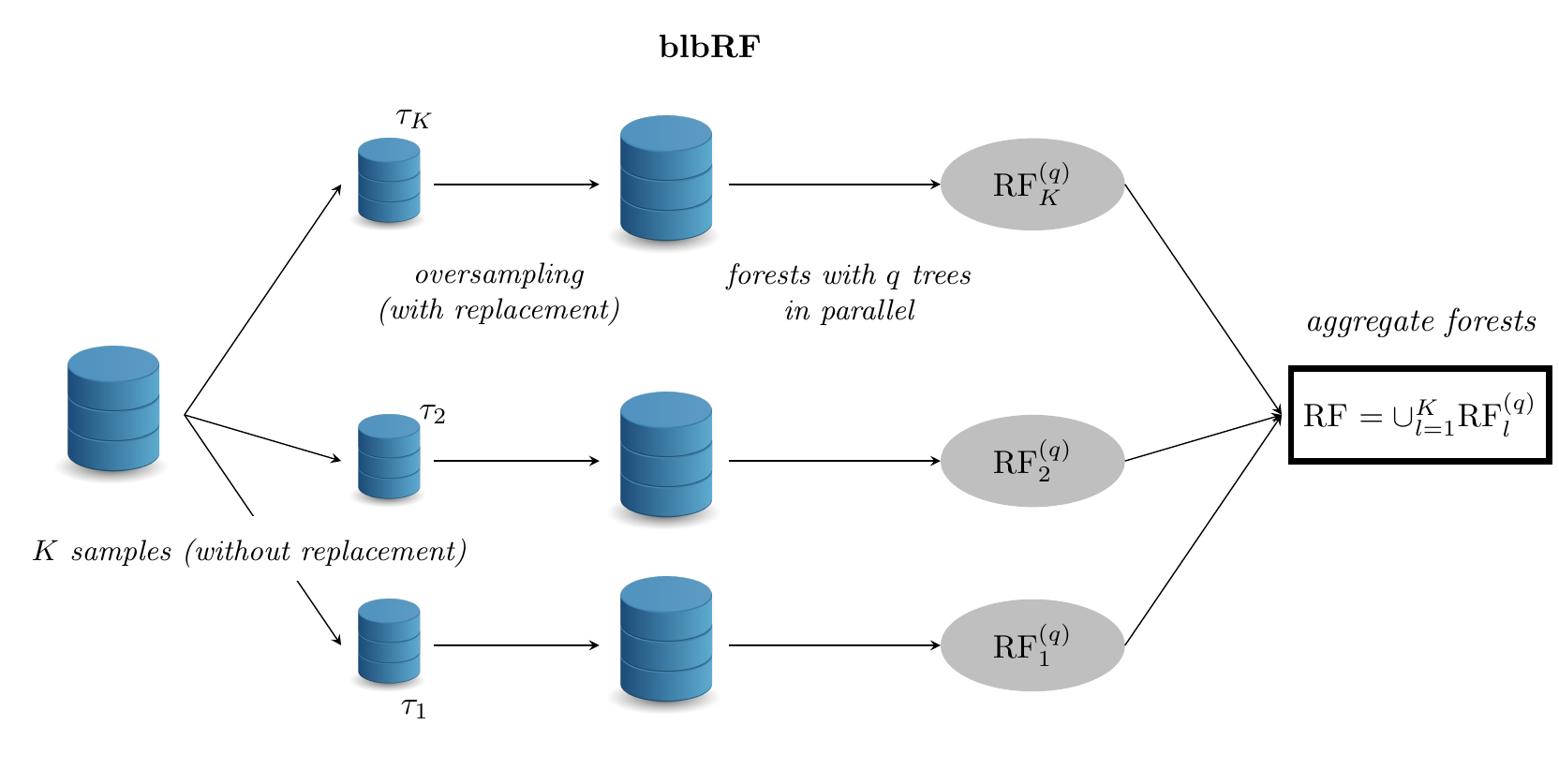}
	\caption{Bag of Little Bootstraps RF ({\bf blbRF}). In this method, a 
subsampling step, performed $K$ times in parallel, is followed by an 
oversampling step which aims at building $q$ trees for each subsample, all 
obtained from a bootstrap sample of size $n$ of the original data. All the 
trees are then gathered into a final forest RF.}
	\label{fig::blbRF}
\end{figure}

More recently, an alternative to $m$-out-of-$n$ bootstrap called 
``Bag of Little Bootstraps'' (BLB) has been described in 
\cite{kleiner_etal_JRSSB2014}. This method aims at building bootstrap samples 
of size $n$, each one containing only $m \ll n$ different observations. The 
size of the bootstrap sample is the classical one ($n$), thus avoiding the 
problem of the bias involved by $m$-out-of-$n$ bootstrap methods. The approach 
is illustrated in Figure~\ref{fig::blbRF}.

It consists in two steps: in a first step, $K$ subsamples, 
$(\tau_l)_{l=1,\ldots,K}$, are obtained, with $m$ observations each, that are 
taken randomly without replacement from the original observations. In a second 
step, each of these subsamples is used to obtain a forest, RF$^{(q)}_l$ with $q 
= \frac{Q}{K}$ trees. But instead of taking bootstrap samples from $\tau_l$, 
the method uses over-sampling and, for all $i \in \tau_l$, computes weights, 
$n_i^l$, from a multinomial distribution with parameters $n$ and $\frac{1}{m} 
\mathbf{1}_m$, where $\mathbf{1}_m$ is a vector with $m$ entries equal to 1.
These weights satisfy $\sum_{i \in \tau_l} n_i^l = n$ and a bootstrap sample of 
the original dataset is thus obtained by using $n_i^l$ times each observation $i$ in  $\tau_l$. 
For each $\tau_l$, $q$ such bootstrap samples are obtained to 
build $q$ trees. These trees are aggregated in a random forest RF$^{(q)}_l$. 
Finally, all these (intermediate) random forests with $q$ trees are gathered 
together in a forest with $Q = qK$ trees. The processing of this method is thus 
simplified by a smart weighting scheme and is manageable even for very large $n$ 
because all bootstrap samples contain only a small number (at most $m$) of 
unique observations from the original data set. The number $m$ is typically of the order 
$n^\gamma$ for $\gamma \in [0.5,1]$, which can be very small compared to the 
typical number of observations (about $0.63n$) of a standard bootstrap sample. 
Interestingly, this approach is well supported by theoretical results because 
the authors of \cite{kleiner_etal_JRSSB2014} prove its equivalence with the 
standard bootstrap method.

\subsubsection{Divide-and-conquer RF (dacRF)}
\label{dac-rf}

\begin{figure}[H]
  \includegraphics[width=\linewidth]{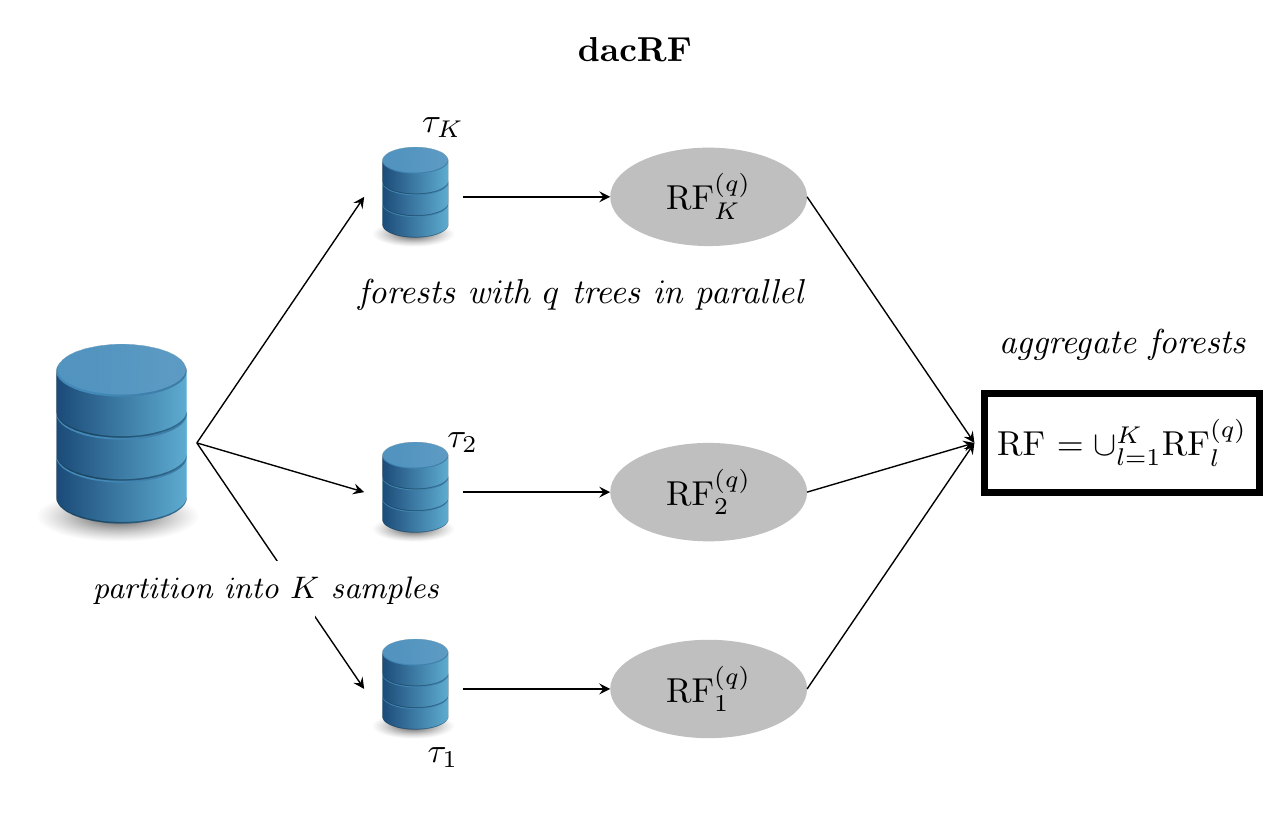}
	\caption{divide-and-conquer RF ({\bf dacRF}). In this method, the original 
dataset is partitionned into $K$ subsets. A random forest with $q$ trees is 
built from each of the subsets and all the forests are finally aggregated in a 
final forest, RF.}
	\label{fig::dacRF}
\end{figure}

Standard alternative to deal with massive datasets while not using subsampling 
is to rely on a ``divide-and-conquer'' strategy. The large problem is divided 
into simpler subproblems and the solutions are aggregated together to solve the 
original problem. The approach is illustrated in Figure~\ref{fig::dacRF}: the 
data are split into small sub-samples, or chunks, of data, $(x_i,y_i)_{i\in 
\tau_l}$, with $\cup_l \tau_l = \{1,\ldots,n\}$ and $\tau_l \cap \tau_{l'} = 
\emptyset$.
Each of these data chunks is processed in parallel and yields to the learning 
of an intermediate RF having a reduced number of trees. Finally, all these 
forests are simply aggregated together to define the final RF.

As indicated in \cite{delrio_etal_IS2014}, this approach is the standard 
MapReduce version of RF, implemented in the Apache$^\textrm{\tiny TM}$ library 
Mahout. MapReduce is a method that proceeds in two steps: in a first step, 
called the Map step, the data set is split into several smaller chunks of data, 
$(x_i,y_i)_{i\in \tau_k}$, with $\cup_k \tau_k = \{1,\ldots,n\}$ and $\tau_k\cap 
\tau_{k'} = \emptyset$, each one being processed by a separate core. These 
different Map jobs are independent and produce a list of couples of the form 
$(\mbox{key}, \mbox{value})$, where ``key'' is a key indexing the data that are 
contained in ``value''. In RF case, the output key is always equal to 1 and the 
output value is the forest learned on the corresponding chunk. Then, in a second 
step, called the Reduce step, each reduce job proceeds all the outputs of the 
Map jobs that correspond to a given key value. This step is skipped in RF case 
since the output of the different Map jobs are simply aggregated together to 
produce the final RF. The MapReduce paradigm takes advantage of the locality of 
data to speed the computation. Each Map job usually processes the data stored in 
a close proximity to its computational unit. As discussed in the next section 
and illustrated in Section~\ref{simulated-biases}, this can yield to biases in 
the resulting RF.

\subsubsection{Mismatches with original RF}
\label{mismatches}

In this section, we want to stress the differences between the previously 
proposed parallel solutions and the original algorithm. Two methods will be 
said ``equivalent'' when they would provide similar results when used on a 
given dataset, up to the randomness in bootstrap sampling. For instance, {\bf 
seqRF} and {\bf parRF} are equivalent since the only difference between the two 
methods are the sequential or parallel learning of the trees.
{\bf sampRF} and {\bf dacRF} are not equivalent to {\bf seqRF} and are both 
strongly dependent on the representativity of the dataset. This is the standard 
issue encountered in survey approaches for {\bf sampRF} but it is also a serious 
limitation to {\bf dacRF} even if this method uses all observations. Indeed, if 
data are thrown in the different chunks with no control on the representativity 
of the subsamples, data chunks might well be specific enough to produce very 
heterogeneous forests: there would be no meaning in simply averaging all those 
trees together to make a global prediction. This is especially an issue 
when using the standard MapReduce paradigm since, as noted by Laptev {\it et 
al.} \cite{laptev_etal_ICVLDB2012}, data are rarely ordered randomly in the Big 
Data world. On the contrary, items are rather clustered on some particular 
attributes are often placed next to each other on disk and the data locality 
property of MapReduce thus leads to very biased data chunks.

Moreover, as pointed out by Kleiner {\it et al.} \cite{kleiner_etal_JRSSB2014}, 
another limit of {\bf sampRF} and {\bf dacRF} but also of {\bf moonRF} comes from 
the fact that each forest is built on a bootstrap sample of size $m$. 
The success of $m$-out-of-$n$ bootstrap samples is highly 
conditioned on the choice of $m$: \cite{bickel_etal_SS1997} reports results 
for $m$ of order $\mathcal{O}(n)$ for successful $m$-out-of-$n$ bootstrap. Bag 
of Little Bootstraps is an appealing alternative since the bootstrap sample 
size is the standard one ($n$). Moreover, \cite{kleiner_etal_JRSSB2014} 
demonstrate a consistency result of the bootstrap estimation in their framework 
for $m = \mathcal{O}(\sqrt{n})$ and $K \sim \frac{n}{m}$ (when $n$ tends to 
$+\infty$). 

In addition, some important features of all these approaches are 
summarized in Table~\ref{table::algo-features}. A desirable property for a high 
computational efficiency is that the number of different observations in 
bootstrap samples is as small as possible.

\begin{table}[H]
	\centering
	\begin{tabular}{l|c|c|c}
		& can be computed & bootstrap & expected nb of\\
		& in parallel & sample size & $\neq$ obs. in \\
		& & & bootstrap samples\\
		\hline
		{\bf seqRF} & yes & $n$ & $0.63n$\\
		{\bf parRF} & ({\bf parRF}) & & \\
		\hline
		{\bf sampRF} & yes but & $m$ & $0.63m$\\
		& not critical & & \\
		\hline
		{\bf moonRF} & yes & $m$ & $m$\\
		\hline
		{\bf blbRF} & yes & $n$ & $m\left[1 - \left(\frac{m-1}{m}\right)^n\right]$\\
		\hline
		{\bf dacRF} & yes & $\frac{n}{K}$ & $0.63\frac{n}{K}$
	\end{tabular}
	\caption{Summary of the main features in the variants of the random forest 
algorithm (excluding online RF, {\bf onRF}).}
	\label{table::algo-features}
\end{table}

\subsubsection{Out-of-bag error and variable importance measure}
\label{big-oob-vi}

OOB error and VI are important diagnostic 
tools to help the user understand the forest accuracy and to perform variable 
selection. However, these quantities may be unavailable directly (or in a 
standard manner) in the RF variants described in the previous sections. This 
comes from the fact that {\bf sampRF}, {\bf moonRF} and {\bf blbRF} use a prior 
subsampling step of $m$ observations. The forest (or the subforests) based on 
this subsample has not a direct access to the remaining $n-m$ observations that 
are always out-of-bag and should, in theory, be considered for OOB computation. 
In general, OOB error (and thus VI) cannot be obtained directly while the forest 
is trained. A similar problem occurs for {\bf dacRF} in which all forests based 
on a given chunk of data are unaware of data the other chunks. In {\bf dacRF}, 
it can even be memory costly to record which data have been used in each chunk 
to obtain OOB afterwards. Moreover, even in the case where this information is 
available, all RF alternatives presented in the previous sections, {\bf 
sampRF}, {\bf moonRF}, {\bf blbRF} and {\bf dacRF}, require to obtain the 
predictions for approximately $n-r m$ OOB observations (with $r=0.63$ for {\bf 
sampRF} and {\bf dacRF}, $r=1$ for {\bf moonRF} and $r=1 - 
\left(\frac{m-1}{m}\right)^n$ for {\bf blbRF}) for all trees, which can be a 
computationally extensive task.

In this section, we present a first approximation of OOB error that can
naturally be designed for {\bf sampRF} and {\bf dacRF}, and a second 
approximation for {\bf moonRF} and {\bf blbRF}. Additional notations used in 
this section are summarized in Table~\ref{table::notations_oob}.
\begin{table}[H]
	\centering
	{\small \begin{tabular}{l|c}
		notation & used for\\
		\hline
		$K$ & number of subsamples\\
		& (equivalent to the number of processes run in parallel here)\\
		$q$ & number of trees in intermediate (smaller) random forests\\
		$\hat{y}_i^l$ & OOB prediction for observation $i \in \tau_l$ by forest 
obtained from $\tau_l$\\
		errForest$^l$ & OOB error of RF$_l^{(q)}$ restricted to $\tau_l$\\
		$\hat{y}_i^{-l}$ & prediction for observation $i \in \tau_l$ by forests 
(RF$^{(q)}_{l'})_{l'\neq l}$\\
		BDerrForest & approximation of OOB in {\bf sampRF}, {\bf blbRF}, {\bf 
moonRF} and {\bf dacRF}
	\end{tabular}}
	\caption{Notations used in Section~\ref{big-oob-vi}.}
	\label{table::notations_oob}
\end{table}

\paragraph{OOB error approximation for {\bf sampRF} and {\bf dacRF}}{
As previously, $(\tau_l)_{l=1,\ldots,K}$ denote the subsamples of data, each of 
size $m$, used to build independent forests in parallel (with $K=1$ for {\bf 
sampRF}). Using each of these samples, a forest with $Q$ ({\bf sampRF}) or $q = 
\frac{Q}{K}$ ({\bf dacRF}) trees is defined, for which an OOB prediction, 
restricted to observations in $\tau_l$, can be calculated: $\hat{y}_i^l$ is 
obtained by a majority vote  on the trees of the forest built from a 
bootstrap sample of $\tau_l$ for which $i$ is OOB.

An approximation of the OOB error of the forest learned from sample $\tau_l$ 
can thus be obtained with $\mbox{errForest}^l = \frac{1}{m} 
\textrm{Card}\left\{i\in\tau_l|y_i\neq \hat{y}_i^l\right\}$. This yields to the 
following approximation of the global OOB error of RF:
\[
	\mbox{BDerrForest} = \frac{1}{n} \sum_{l=1}^K m \times \mbox{errForest}^l
\]
for {\bf dacRF} or simply BDerrForest $ = \mbox{errForest}^1$ for {\bf sampRF}.
}

\paragraph{OOB error approximation for {\bf moonRF} and {\bf blbRF}}{
For {\bf moonRF}, since samples are obtained without replacement, there are
no OOB observations associated to a tree. However we can compute an 
OOB error as in standard forests, restricted to the set $\cup_{l=1}^Q \tau_l$ 
of observations that have been sampled in at least one of the subsamples 
$\tau_l$. This leads to obtain an approximation of the OOB error, BDerrForest, 
based on the prediction of approximately $(Q-1)m$ observations (up to the few 
observations that belong to several subsamples, which is very small if $m \ll 
n$) that are OOB for each of the $Q$ trees. This corresponds to an important 
computational gain as compared to the standard OOB error that would have 
required the prediction of approximately $n-m$ observations for each tree. 

For {\bf blbRF}, a similar OOB error approximation can be computed using 
$\cup_{l=1}^K \tau_l$. Indeed, since trees are built on samples of size $n$ 
obtained with replacement from $\tau_l$ (having a size equal to $m$), and again 
provided that $m \ll n$, there are no OOB observations associated to the trees 
with high probability. Again assuming that no observation belong to several 
subsamples $\tau_l$, the OOB prediction of an observation in $\tau_l$ can be 
approximated by a majority vote law based on the predictions made by subforests 
(RF$^{(q)}_{l'})_{l' \neq l}$. If this prediction is denoted by 
$\hat{y}_i^{-l}$, then the following approximation of the OOB error can be 
derived:
\[
	\mbox{BDerrForest} = \frac{1}{Km} \sum_{l=1}^K \mathrm{Card} 
\left\{i\in\tau_l \: | \: y_i \neq \hat{y_i}^{-l} \right\}.
\]
Again, for each tree, the number of predictions to make to compute this error 
is $(K-1)m$, which is small compared to the $n-m$ predictions that would have 
been performed to compute the standard OOB error.
}

Similar approximations can also be defined for VI (not investigated in this 
paper for the sake of simplicity).

\subsection{Online random forests}
\label{online-rf}

The general idea of online RF ({\bf onRF}), introduced by Saffari 
{\it et al.} \cite{saffari_etal_ICCV2009}, is to adapt RF methodology, in order 
to handle the case where data arrive sequentially. An online framework supposes 
that at a given time step one does not have access to all the data from the 
past, but only to the current observation. {\bf onRF} are first defined in 
\cite{saffari_etal_ICCV2009} and detailed only for classification problems. They 
combine the idea of online bagging, also called Poisson bootstrap, from 
\cite{oza_russel_IWAIS2001,lee_clyde_JMLR2004,hanley_macgibbon_CMPB2006}, 
Extremely Randomized Trees (ERT) from \cite{geurts_etal_ML2006}, and a mechanism 
to update the forest each time a new observation arrives.

More precisely, when a new data arrives, the online bagging updates $k$ times a 
given tree, where $k$ is sampled from a Poisson distribution to mimic a batch 
bootstrap sampling. This means that this new data will appear $k$ times in the 
tree, which mimics the fact that one data can be drawn $k$ times in the batch 
sampling (with replacement). ERT is used instead of original Breiman's RF, 
because it allows for a faster update of the forest: in ERT, $S$ splits ({\it 
i.e.}, a split variable \textit{and} a split value) are randomly drawn 
for every node, and the final split is optimized only among those $S$ candidate 
splits. Moreover, all decisions given by a tree are only based on the 
proportions of each class label among observations in a node. {\bf onRF} keep 
up-to-date (in an online manner) an heterogeneity measure based on these 
proportions, used to determine the class label of a node. So when a node is 
created, $S$ candidate splits (hence $2S$ candidate new nodes) are randomly 
drawn and when a new data arrives in an existing node, this measure is
updated for all those $2S$ candidate nodes. This mechanism is repeated until a 
stopping condition is realized and the final split minimizes the heterogeneity 
measure among the $S$ candidate splits. Then a new node is created and so on.

From the theoretical viewpoint, the recent article \cite{denil_etal_ICML2013} 
introduces a new variant of {\bf onRF}. The two main differences with the 
original {\bf onRF} are that, 1) no online bootstrap is performed. 2)  Each 
point is assigned to one of two possible streams at random with fixed 
probability. The data stream is then randomly partitioned in two streams: 
the structure stream and the estimation stream. Data from structure stream only 
participate on the splits optimization, while data from estimation stream are 
only used to allocate a class label to a node. Thanks to this partition, the 
authors manage to obtain consistency results of {\bf onRF}.

\cite{saffari_etal_ICCV2009} also describes an online estimation of the OOB 
error: since a given observation is OOB for all trees for which the Poisson 
random variable used to replicate the observation in the tree is equal to 0, 
the prediction provided for such a tree $t$ is used to update 
$\mbox{errTree}_t$. However, since the prediction cannot be re-evaluated after 
the tree has been updated with next data, this approach is only an approximation 
of the original $\mbox{errTree}_t$. Moreover, as far as we know, this 
approximation is not implemented in the python library RFTK 
\footnote{\url{https://github.com/david-matheson/rftk}} which provides an 
implementation of {\bf onRF} used in experiments of 
Section~\ref{simu-online-rf}. Finally, since permuting the values of a given 
variable when the observations are processed online and are not stored after 
they have been processed is still an open issue for which 
\cite{saffari_etal_ICCV2009,denil_etal_ICML2013} give no solution. Hence, VI 
cannot be simply defined in this framework. 

\section{Experiments}
\label{experiments}

The present section is devoted to numerical experiments on a massive simulated 
dataset (15 millions of observations) as well as a real world dataset
(120 millions of observations), which aim at illustrating and comparing 
the five variants of RF for Big Data introduced in Section~\ref{scaling-rf-bd}.
The experimental framework and the data simulation
model are first presented. Then four variants involving parallel
implementations of RF are compared, and online
RF is also considered. A specific focus on the influence of 
biases in subsampling and splitting is performed. Finally, we analyze the performance 
obtained on a well-known real-world benchmark for Big Data experiments that 
contains airline on-time performance data.

\subsection{Experimental framework and simulation model}
\label{exp-framework}

All experiments have been conducted on the same server (with concurrent 
access), with 8 processors AMD Opteron 8384 2.7Ghz, with 4 cores each, a total 
RAM equal to 256 Go and running on Debian 8 Jessie. Parallel methods
were all run with 10 cores.

There are strong reasons to carry out 
experimentations in a unified way involving codes in \RR{}. This will be the 
case in this section except for {\bf onRF} in Section~\ref{simu-online-rf}.
Due to their interest, {\bf onRF} are considered in  experimental part of the 
paper, even if, due to the lack of available program implemented in \RR{},
an exception has been made using a python code.
To allow fair comparisons between the other methods and to make them
independent from a particular software 
framework or a particular programming language, all methods have been 
programmed using the following packages:
\begin{itemize}
	\item the package 
\pkg{readr} \cite{wickham_francois_R2015} (version 0.1.1), which allows to 
read more efficiently flat and tabular text files from disk;
	\item the package \pkg{randomForest} \cite{liaw_wiener_RN2002} (version 
4.6-10), which implements RF algorithm using Breiman and Cutler's original 
Fortran code;
	\item the package \pkg{parallel} \cite{RCT_R2016} (version 3.2.0), which is 
part of \RR{} and supports parallel computation.
\end{itemize}

To address all these issues, simulated data are studied in this section. They 
correspond to a well controlled model and can thus be used to obtain 
comprehensive results on the various questions described above. The simulated 
dataset corresponds to 15,000,000 observations generated from the model 
described in \cite{weston_etal_JMLR2003}: this model is an equiprobable two 
class problem in which the variable to predict, $Y$, takes values in $\{-1,1\}$ 
and the predictors are, for 6 of them, true predictors, whereas the other ones 
(in our case only one) are random noise. The simulation model is defined 
through the 
law of $Y$ (${P}(Y=1) = {P}(Y=-1) = 0.5$) and the conditional distribution of 
the $(X^j)_{j=1,\ldots,7}$ given $Y=y$:
\begin{itemize}
	\item with probability equal to 0.7, $X^j \sim \mathcal{N}(jy,1)$ for 
$j\in\{1,2,3\}$ and $X^j \sim \mathcal{N}(0,1)$ for $j\in\{4,5,6\}$ (submodel 1);
	\item with probability equal to 0.3, $X^j \sim \mathcal{N}(0,1)$ for 
$j\in\{1,2,3\}$ and $X^j \sim \mathcal{N}((j-3)y,1)$ for $j\in\{4,5,6\}$ (submodel 2);
	\item $X^7\sim \mathcal{N}(0,1)$.
\end{itemize}
All variables are centered and scaled to unit variance after the simulation 
process, which gave a dataset which size (in plain text format) was equal to 
$1.9$ Go. Compared to the size of available RAM, this dataset was relatively 
moderate which allowed us to perform extensive comparisons while being in the 
realistic Big Data framework with a large number of observations.

This 15,000,000 observations of this dataset were first randomly ordered.
Then, to illustrate the effect of representativeness of data in different
sub-samples in both divide-and-coquer and online approaches, two permuted 
versions of this same dataset were considered (see 
Figure~\ref{fig::unbalanced_xbias} for an illustration):
\begin{figure}[H]
  \includegraphics[width=\linewidth]{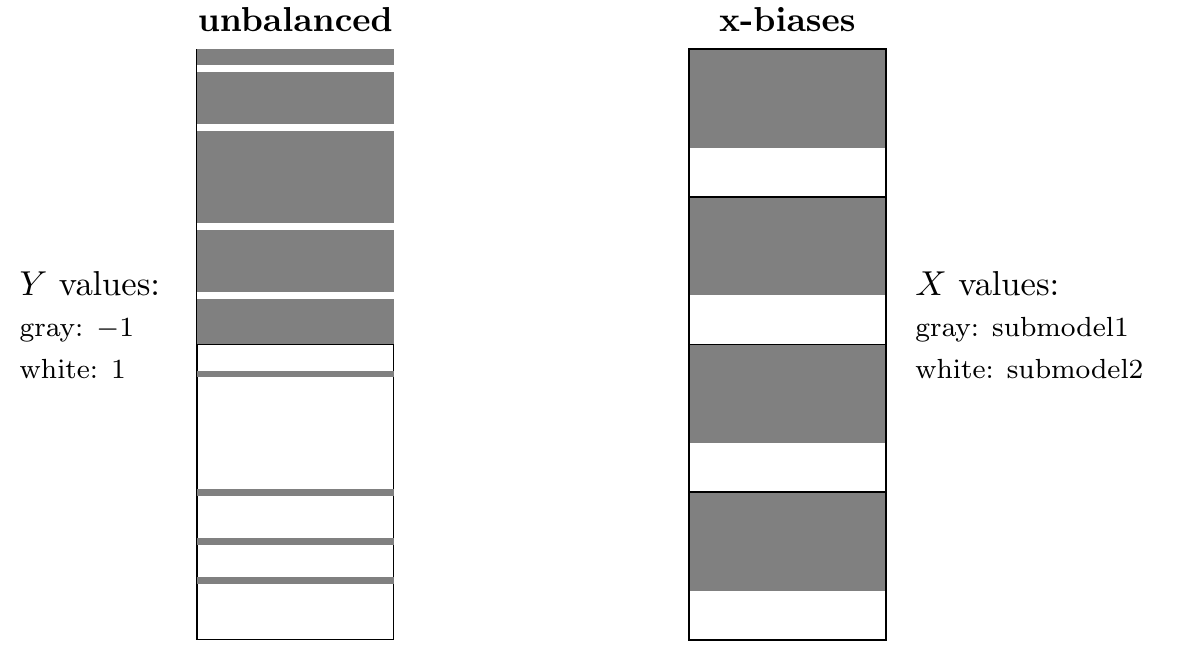}
	\caption{Illustration of the datasets {\bf unbalanced} (left) and {\bf 
x-biases} (right)}
	\label{fig::unbalanced_xbias}
\end{figure}
\begin{itemize}
	\item \textbf{unbalanced} will refer to a permuted dataset
	in which $Y$ values arrive with a particular pattern.
	More precisely, we permuted the
	observations so that the first half of the observations contain a proportion 
$p$ (with $p\in\{10;1\}$\%) of observations
	coming from the first class ($Y=1$), and the other half contains the same 
proportion of observations from the second class
	($Y=-1$);
	\item \textbf{x-biases} will refer to a permuted dataset
	in which $X$ values arrive with a particular pattern.
	More precisely, in that case, the data are split into $P$ parts in which the 
first 70\% of the observations are coming from submodel 1 
	and the last 30\% are coming from 
submodel 2.
\end{itemize}

\subsection{Four RF methods for Big Data involving parallel implementations}
\label{simulated}

The aims of the simulations of this subsection were multiple: firstly,
different approaches 
designed to handle Big Data with RF were compared. The comparison was made on 
the point of view of the computational effort needed to train the classifier 
and also in term of its accuracy. Secondly, the differences between the OOB 
error estimated by standard methods corresponding to a given approach (which 
generally uses only a part of the data to be computed) was compared to the OOB 
error of the classifier estimated on the whole data set.

All along this subsection we use a simulated dataset corresponding to
15,000,000 observations generated from the model described in
Section~\ref{exp-framework} and randomly ordered.
With the \pkg{readr} package, loading this dataset
took approximately one minute.

As a baseline for comparison, a standard RF with 100 trees was trained
in a sequential way with the \RR{} package \pkg{randomForest}. This package 
allows to control the complexity of the trees in the forest by setting a 
maximum number of terminal nodes (leaves). By default, fully developed trees 
are grown, with unlimited number of leaves, until all leaves are pure ({\it 
i.e.} composed of observations all belonging to the same class). Considering the
very large number of observations, the number of leaves was limited to
500 in our experiments. The training of this forest took approximately $7$ 
hours and the resulting
OOB error was equal to $4.564e^{-3}$ and has served as a baseline for the other 
experiments.

\begin{figure}
\centering
	\includegraphics[width=\linewidth]{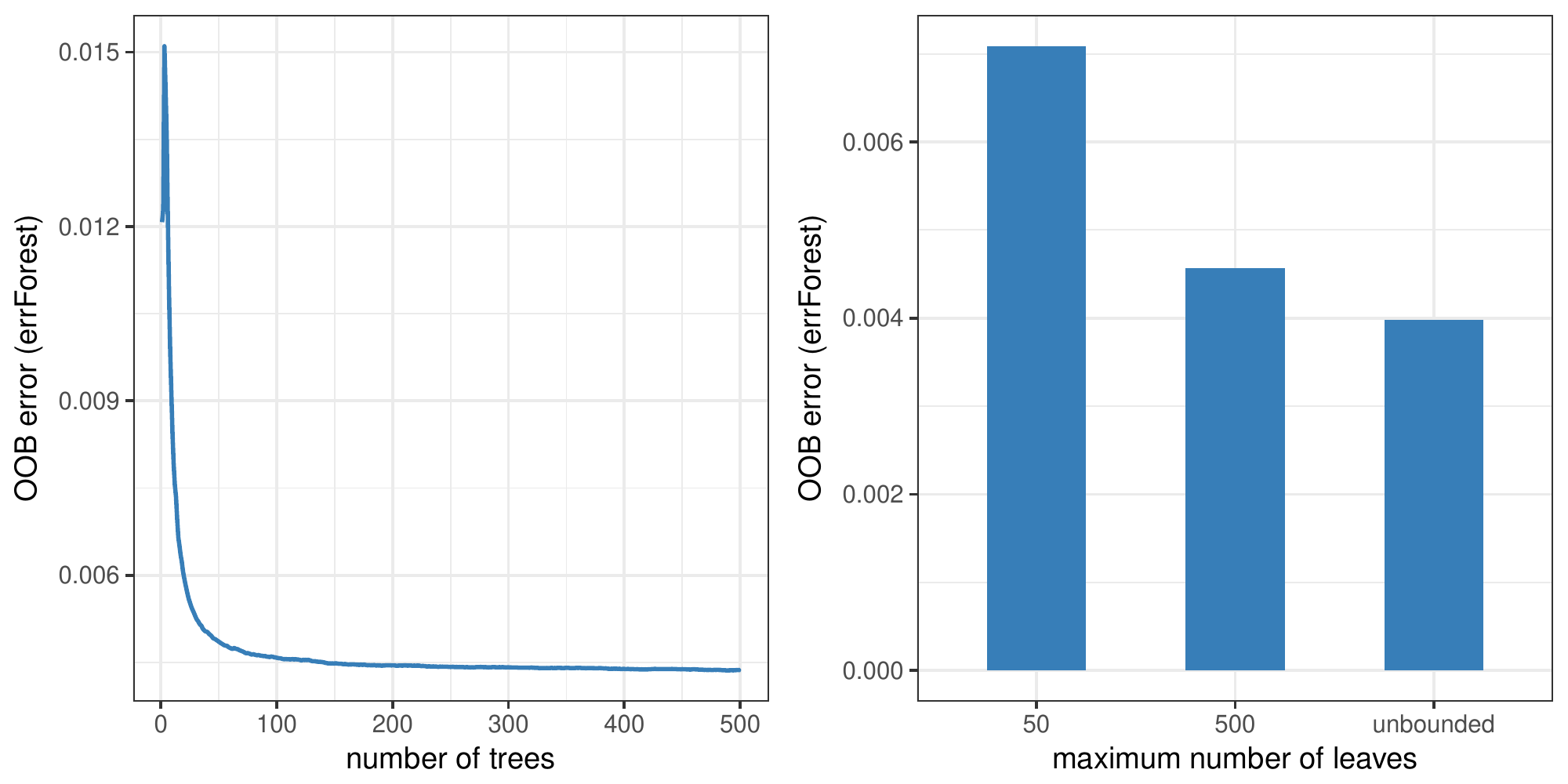}
	\caption{OOB error evolution for {\bf seqRF} versus the
	number of trees (left), and the maximum number of leaves (right).}
		\label{fig::RFseq}
\end{figure}

As illustrated by the left-hand side of Figure~\ref{fig::RFseq},
the OOB error of {\bf seqRF}
(with a total number of trees equal to 500) stabilizes between $100$ and $200$ 
trees.
The training of the RF with $500$ trees took approximately $18$ hours.
Hence, we chose to keep a limited number of trees of 100, which seems a good
compromise between accuracy and computational time.
The choice of a maximum number of leaves of $500$ was also 
motivated by
the fact that maximal trees did not bring much improvement in accuracy, as
shown in the right-hand side of Figure~\ref{fig::RFseq}. On the contrary,
it increases the final RF complexity significantly (maximal trees contain
approximately 60,000 terminal nodes).

We designed experiments to compare this sequential forest ({\bf seqRF}) to the
four variants introduced in Section~\ref{scaling-rf-bd}, namely: {\bf sampRF},
{\bf moonRF}, {\bf blbRF} and {\bf dacRF} (see Table~\ref{table::method-names}
for definitions). In this section, the purpose is only to compare the methods 
themselves so all subsamplings were done in such a way that the subsamples
were representative of the whole 
dataset from the $X$ and $Y$ distributional viewpoint. 

The different results are compared through the computational time needed by 
every method (real elapsed time as returned by \RR) and the prediction 
performance. This last quantity was assessed in three ways:
\begin{itemize}
	\item[{\it i)}] errForest, which is defined in Equation~(\ref{def-errforest} )
and
	refers to the standard OOB error of a RF. 
	This quantity is hard to obtain with the different 
methods described in this chapter when the sample size is large but we 
nevertheless computed it to check if the approximations usually used to 
estimate this quantity are reliable;
	\item[{\it ii)}] BDerrForest, which is the approximation of errForest
	defined in Section~\ref{big-oob-vi};
	\item[{\it iii)}] errTest, which is a standard test error using a test sample,
	with 150,000 observations, generated independently from the training sample.
\end{itemize}
In all simulations, the maximum number of leaves in the trees was set to 500.
In addition, errOOB and errTest were found always indistinguishable,
which confirms that OOB error is a good estimation of the prediction error.

\begin{figure}
\centering
	\includegraphics[width=\linewidth]{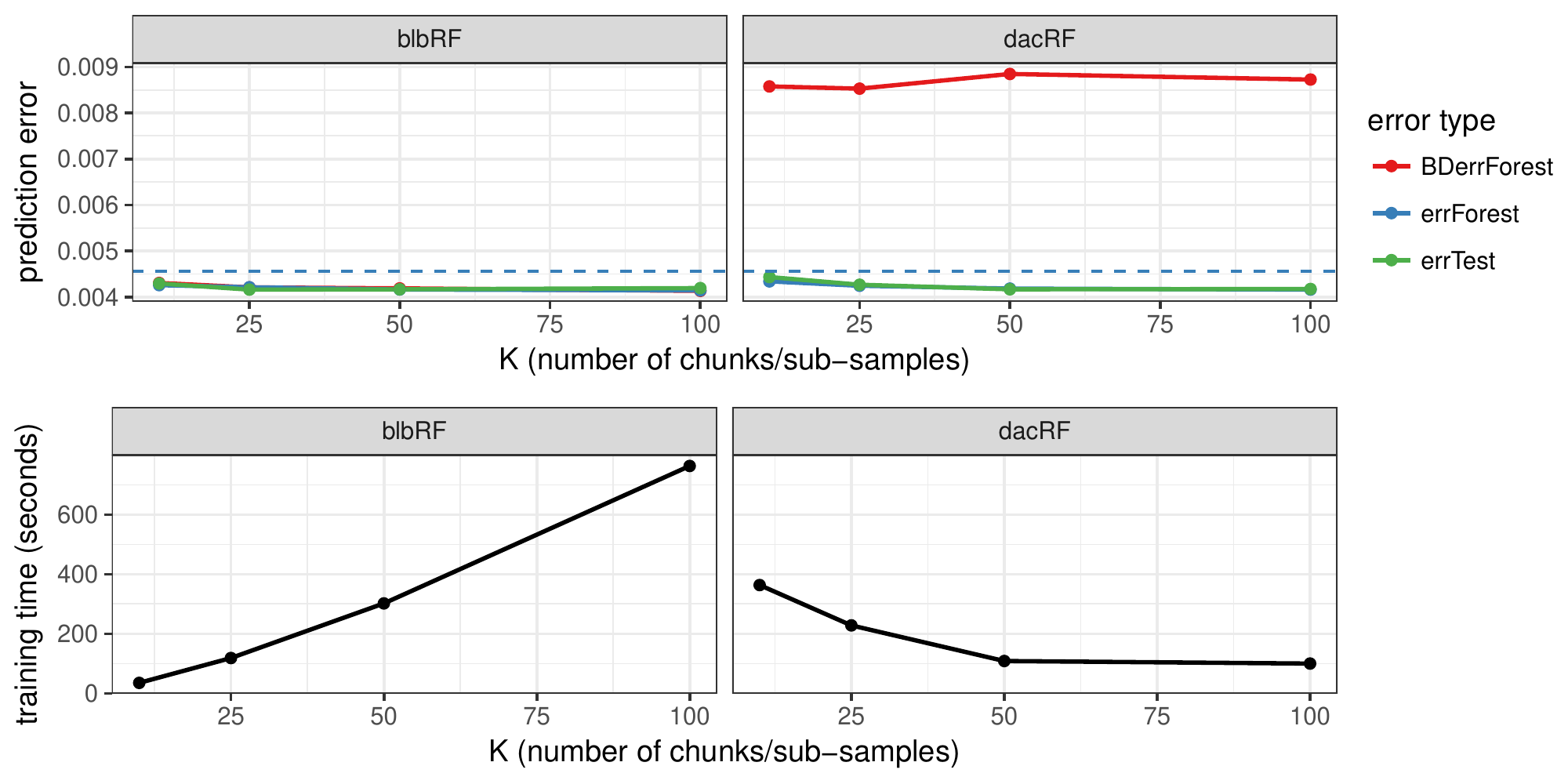}
	\caption{Evolution of prediction error (top) and 
	computational time for training (bottom) versus $K$. $K$ is the number of 
chunks
	for {\bf dacRF} (right) or the number of sub-samples for {\bf blbRF}
	(left). The number of trees, $q$, is set to $10$.}
	\label{fig::blb_mrrf_nbchunks}
\end{figure}

First, the impact of $K$ and $q$ for {\bf blbRF} and {\bf
dacRF} was studied. As shown in Figure~\ref{fig::blb_mrrf_nbchunks}, when
$q$ is set to $10$, {\bf blbRF} and {\bf dacRF} are quite insensitive to the
choice of $K$. However, BDerrForest is a very pessimistic approximation of the 
prediction error for
{\bf dacRF}, whereas it gives good approximations for {\bf blbRF}.
Computational time for training is obviously linearly increasing for {\bf 
blbRF},
as we built more sub-samples, whereas it is decreasing for {\bf dacRF}, because
the size of each chunk becomes smaller.

\begin{figure}
\centering
	\includegraphics[width=\linewidth]{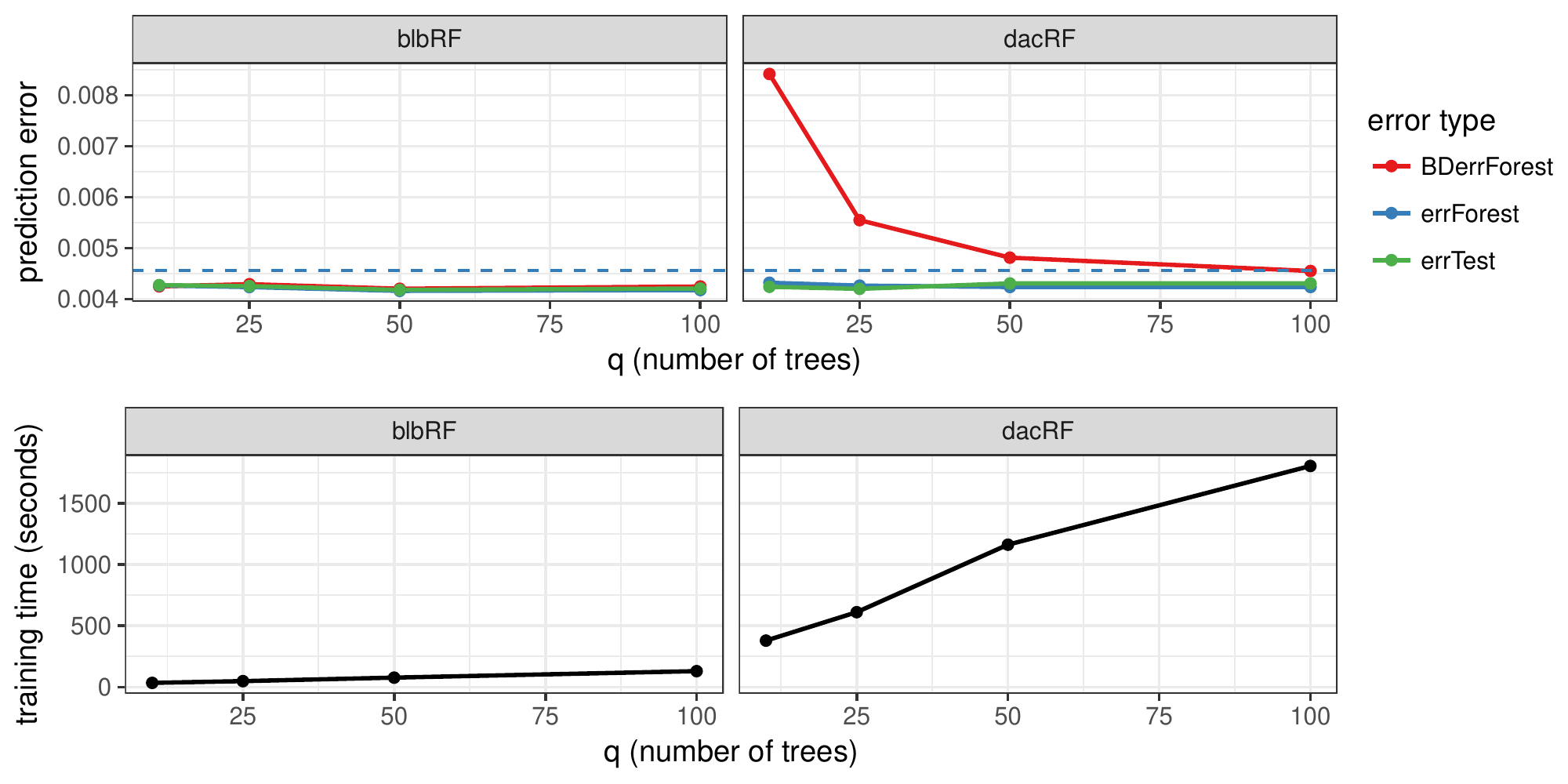}
	\caption{Evolution of the prediction error (top) and 
	computational time for training (bottom) versus $q$. $q$ is the 
number of trees
	in each chunk for {\bf dacRF} (right) or the number of trees in each 
sub-sample for
	{\bf blbRF} (left). $K$ is set to $10$.}
	\label{fig::blb_mrrf_nbtrees}
\end{figure}

Symmetrically, $K$ was then fixed to $10$ to illustrate the effect of the 
number of trees in each
chunk/sub-samples. Results are provided in Figure~\ref{fig::blb_mrrf_nbtrees}. 
Again, {\bf blbRF} is
quite robust to the choice of $q$. On the contrary, for {\bf dacRF}, the
number of trees built in each chunk must be quite high to get an unbiased
BDerrForest, at a cost of a substantially increased computational time.
In other simulations for {\bf dacRF}, $q$ was also set to $100$ and $K$ was 
increased but this did not give any improvement (not shown).
Due to these conclusions, the values $K=10$ and $q=50$ were chosen for 
{\bf blbRF} and the values $K=10$, $q=100$ were chosen for {\bf dacRF} in the 
rest of the simulations.

\begin{figure}
\centering
	\includegraphics[width=\linewidth]{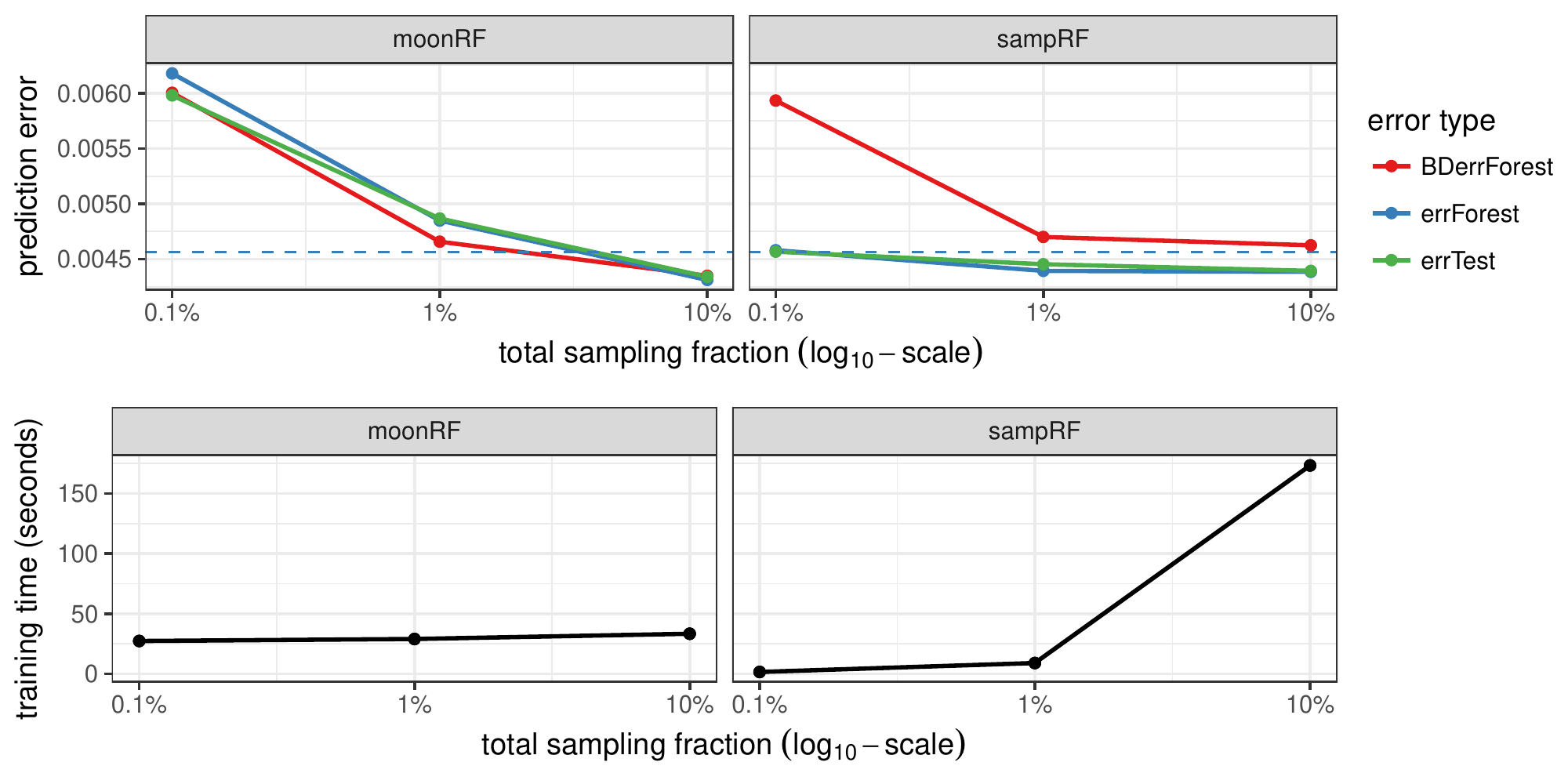}
	\caption{Evolution of the prediction error (top) and computational
	time for training (bottom) versus the sampling fraction
	($\log_{10}$-scale) used in {\bf moonRF} (left) and {\bf sampRF}
	(right). The number of trees is set to $100$.}
	\label{fig::moonrf_samprf}
\end{figure}

Second, the impact of the sampling fraction, $f = \frac{m}{n}$ was studied for
{\bf sampRF} and {\bf moonRF}, with a number of trees set to $100$.
More precisely, for {\bf sampRF},
a subsample containing $m$ observations was randomly drawn for the entire 
dataset, with
$f \in \{0.1, 1, 10\}$\%. Results
(see the right-hand side of Figure~\ref{fig::moonrf_samprf}) show
that BDerrForest is quite unbiased as soon as $f$ is larger than
1\%. Furthermore, $f=10$\% leads to some increase in computational
time needed for training, even if this time is around $10$ times smaller than 
the one needed to train {\bf dacRF} with $10$ chunks and $100$ trees.
For {\bf moonRF}, as the $100$ trees are built on samples with $m$ different
observations 
each, the sampling fraction was varied in $\{10^{-5}, 10^{-4}, 10^{-3}\}$,
in order to get a fraction of observations used by the entire forest (total 
sampling fraction, represented on the $x$-axes of the figure)
comparable to the one used in {\bf sampRF}. The left-hand part of
Figure~\ref{fig::moonrf_samprf} shows that
BDerrForest gives quite unbiased estimations of the prediction error.
Moreover, the computational time for training remains low. The increase
of the prediction error when $f=0.1$\% is explained by the fact that subsamples 
contain only $150$ observations in this case. Based on these experiments, the 
total sampling fraction was set to 1\% for both {\bf sampRF} and
{\bf moonRF} in the rest of the simulations.

Several conclusions can be driven from these results. First, the computational 
time needed to train all these Big Data versions of RF is almost the same and 
quite reduced (about a few minutes) compared to the sequential RF. The fastest 
approach is to extract a very small subsample and the slowest is the {\bf dacRF}
approach with 10 chunks of 100 trees each (because the number of observations 
sent to each chunk is not much reduced compared to the original dataset). The 
results are not shown for the sake of simplicity but the performances are also 
quite stable: when a method was trained several times with the same parameters, 
the performances were almost always very close. 

Regarding the errors, it has first to be noted that the prediction error 
(as assessed with errTest) is much better estimated by errForest than by the 
proxy of the OOB error provided by BDerrForest. In 
particular, BDerrForest tends to be biased for {\bf sampRF} and {\bf moonRF}
approaches when the fraction of samples is very small and it tends to
overestimate the prediction error (sometimes strongly) for {\bf dacRF}. 

Finally, many methods achieve a performance which is quite close to that of the 
standard RF algorithm: {\bf sampRF} and {\bf moonRF} approaches are quite close
to the standard algorithm {\bf seqRF}, even for very small subsamples
(with at least 0.1\% of the original 
observations, the difference between the two predictors is not very important). 
{\bf blbRF} is also quite close to {\bf seqRF} and remarkably 
stable to a change in its parameters $K$ and $q$. Finally, {\bf dacRF} also
gives an accurate predictor but its BDerrForest error estimation is close to the
prediction error only when the number of trees in the forest is
large enough: this is obtained at the price of a higher
computational cost (about 10 times larger than for the other approaches).

\subsection{More about subsampling biases and tree depth}
\label{simulated-biases}

In the previous section, simulations were conducted with representative 
subsamples and a maximum number of leaves equal to 500 for every tree 
in every forest. The present section pushes the analysis a bit further by 
specifically investigating the influence of these two features on the results. 
All simulations were performed with the same dataset and the same computing 
environment than in the previous section. Finally, the different
parameters for the RF methods were fixed in light of the previous section: {\bf 
blbRF} and
{\bf dacRF} were learned respectively with $K=10$ and $q=50$ and with $K=10$, 
$q=100$,
whereas {\bf moonRF} and {\bf sampRF} were learned with total sampling 
fraction equal to $0.1$\%.

As explained in Section~\ref{parallel-rf}, {\bf dacRF} can be influenced by the
lack of representativity of the data sent to the different chunks.
In this section, we evaluate the influence of such cases in two different
directions. We have considered the non representativity of observations
in the different chunks/sub-samples, firstly according to $Y$ values using the
{\bf unbalanced} dataset and secondly, according to $X$ values using the
{\bf x-biases} dataset (see
Section~\ref{exp-framework} for a description of these two datasets).
For {\bf dacRF}, this simulation corresponds to the case where the subforests
built from the different chunks are very heterogeneous. This issue has been 
discussed in
Section~\ref{mismatches} and we will show that it indeed has a strong impact in
practice.

\begin{figure}
\centering
	\includegraphics[width=\linewidth]{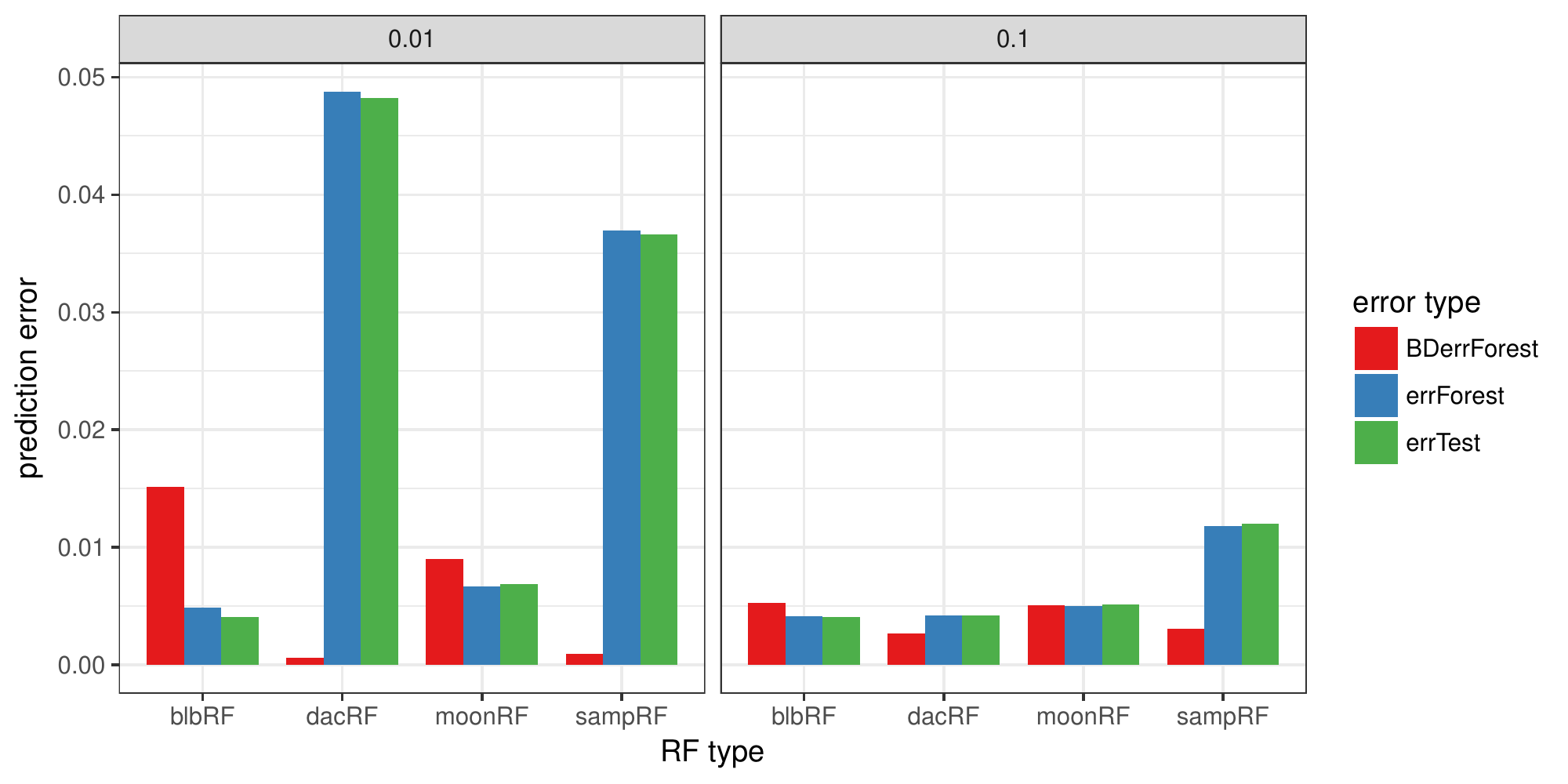}
	\caption{Prediction error behavior for 4 RF methods for {\bf unbalanced}
	data. Unbalanced proportion $p$ is set to 0.01 (left) or to 0.1 (right).}
	\label{fig::unbal}
\end{figure}

Results associated to the {\bf unbalanced} case are presented in
Figure~\ref{fig::unbal}. In this case, data are organized so that, for
{\bf dacRF}, half of the chunks have a proportion $p \in \{0.01, 0.1\}$
of observations from the first class ($Y=1$), and the other half have the same 
proportion of observations from the
second class ($Y=-1$).
For {\bf blbRF} and {\bf moonRF}, half of the sub-samples were drawn in order
to get a proportion $p$ of observation from the first class and the other half 
the same proportion of observations from
the second class.
Finally, as there is only one subsample to draw for {\bf sampRF}, it has
been obtained with a proportion $p$ of observations of the first class. Hence,
the results associated to {\bf sampRF} are not fully comparable to the other 
two.

The first fact worth noting in these results is again that errOOB and errTest
are always very close, whereas BDerrForest is more and more biased as
$p$ decreases. For $p=0.1$, BDerrForest bias is rather stable for all
methods, except for {\bf sampRF} (which is explained by the fact that only one 
subsample is chosen and thus 90\% of the observations are coming from
the second class). When $p=0.01$ (which corresponds to a quite extreme
situation), we can see that {\bf dacRF} is the most affected method, in
terms of BDerrForest (BDerrForest strongly underestimates the prediction error) 
but also in
terms of errOOB and errTest because these two quantities increase a lot.

Interestingly, {\bf moonRF}
is quite robust to this situation, whereas {\bf blbRF} has a BDerrForest
which strongly overestimates the prediction error. The difference
of behavior between these two last methods might from the fact that, in
our setting, 100 sub-samples are drawn for {\bf moonRF} but only 10 for {\bf
blbRF}.

\begin{figure}
\centering
	\includegraphics[width=\linewidth]{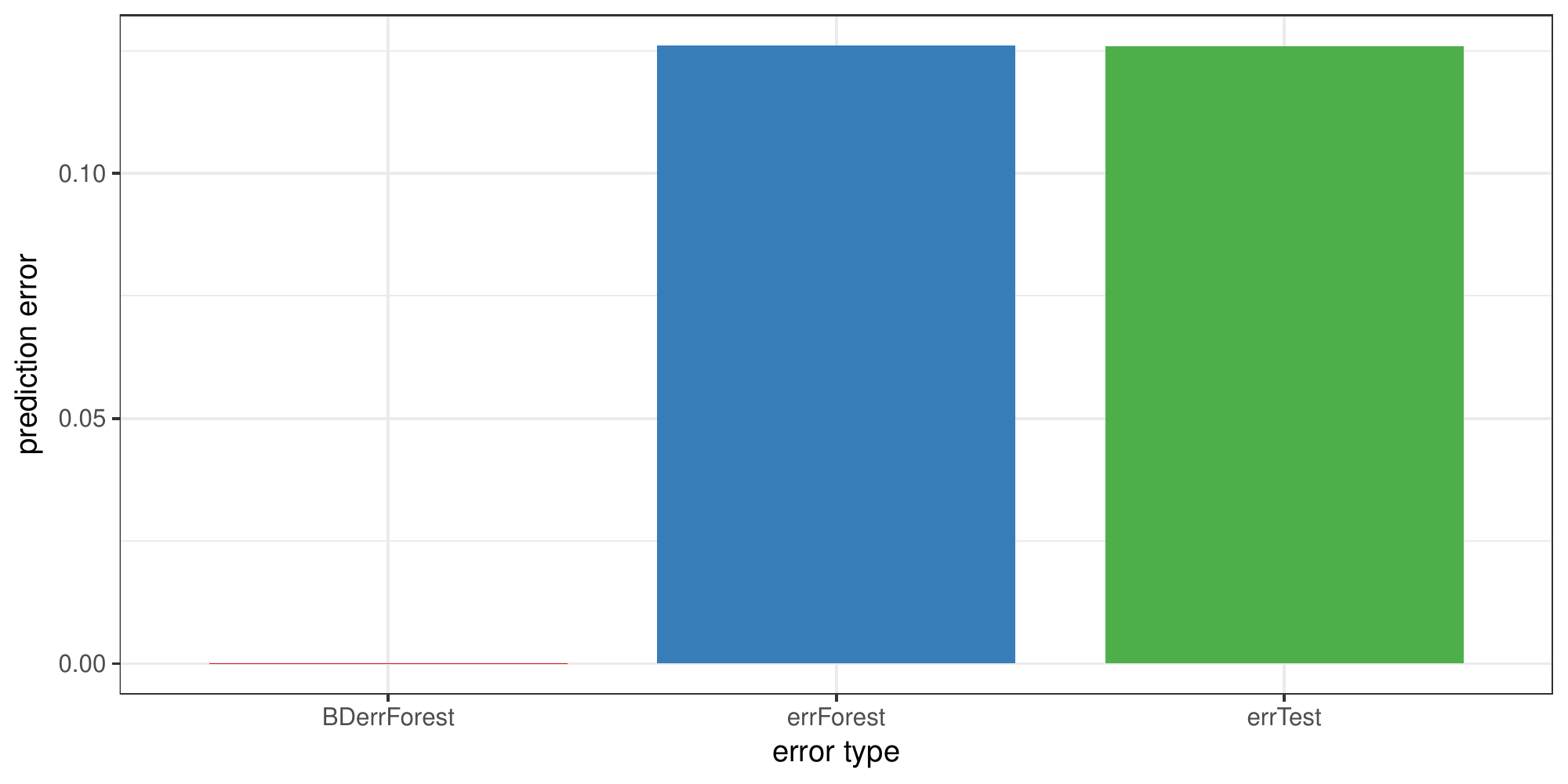}
	\caption{Prediction errors for {\bf x-biases} with {\bf dacRF} ($K=10$ and 
$q=100$).}
	\label{fig::xbias}
\end{figure}

A similar conclusion is obtained for biases towards $X$ values: simulations 
have been performed for {\bf dacRF} with {\bf x-biases} obtained by 
partitionning the data into 2 parts (as illustrated on the right-hand side of 
Figure~\ref{fig::unbalanced_xbias}), leading to 7/10 of the $K=10$ chunks of 
data to contain only observations from submodel 1 and the other 3/10 chunks 
contaning only observations from submodel 2. Results are given in 
Figure~\ref{fig::xbias}. 
This result shows that the performance
of the forest is strongly deteriorated when subforests are based on 
observations coming from different distributions $X|Y$: in this case, the test 
misclassification rate is
multiplied by a factor of more than 50. Moreover, BDerrForest appears to be a 
very bad estimation of the prediction error of the forest.

Finally, the issue of tree depth is investigated more closely.
As mentioned above, the maximum number of leaves 
was set to 500 in order to get comparable tree complexities. However 
homogeneity (in terms of classes) of leaves differs when a tree is built on the 
entire dataset or on a fraction of it. To illustrate this, the mean Gini index
(over all leaves of a tree and over 100 trees) was computed (it is defined by 
$2 
\hat{p} (1- \hat{p})$, with $\hat{p}$ the proportion of observations of class 
1 in a leaf). Results are reported in Table~\ref{table::nbleaves}.

\begin{table}
	\centering
	\fbox{\begin{tabular}{l|c|c|c|c}
		Sampling & Comp. & Max. tree & Pruned tree & mean Gini\\
		fraction & time & size & size & \\
		\hline
		{\bf 100\%} & 5 hours & 60683 & 3789 & 0.233 \\
		{\bf 10\%}  & 13 min & 6999 & 966 & 0.183 \\
		{\bf 1\%}   & 23 sec & 906 & 187 & 0.073 \\
		{\bf 0.1\%} & 0.01 sec & 35 & 10 & 0.000
	\end{tabular}}
	\caption{Number of leaves and leaves heterogeneity of trees built
	on various fractions of data. Second column indicates computational
	time needed to built one tree, while number of leaves of the maximal tree
	and the optimal pruned tree are given in third and fourth column respectively.
	The last column the mean Gini index over all leaves of a tree and over 100 
trees.}
	\label{table::nbleaves}
\end{table}

For sampling fractions equal to 0.1\% or 1\%, tree leaves are pure ({\it i.e.}, 
contain
observations from only one class). But for sampling fractions equal to 100\% 
and 
10\%, the heterogeneity of the leaves is more 
important. The effect of trees depth on RF performance was thus investigated.
Recall that in RF all trees are typically grown to maximal trees (splits are
performed until each leaf is pure) and that in CART an optimal tree is obtained 
by pruning the maximal tree. Table~\ref{table::nbleaves} contains the number of 
leaves of the maximal tree and the optimal CART tree associated to each 
sampling fraction. Trees with 500 leaves are very far from maximal trees in 
most 
cases and even far from optimal CART tree for sampling fractions equal to 100\% 
and 10\%.

\begin{figure}
\centering
	\includegraphics[width=\linewidth]{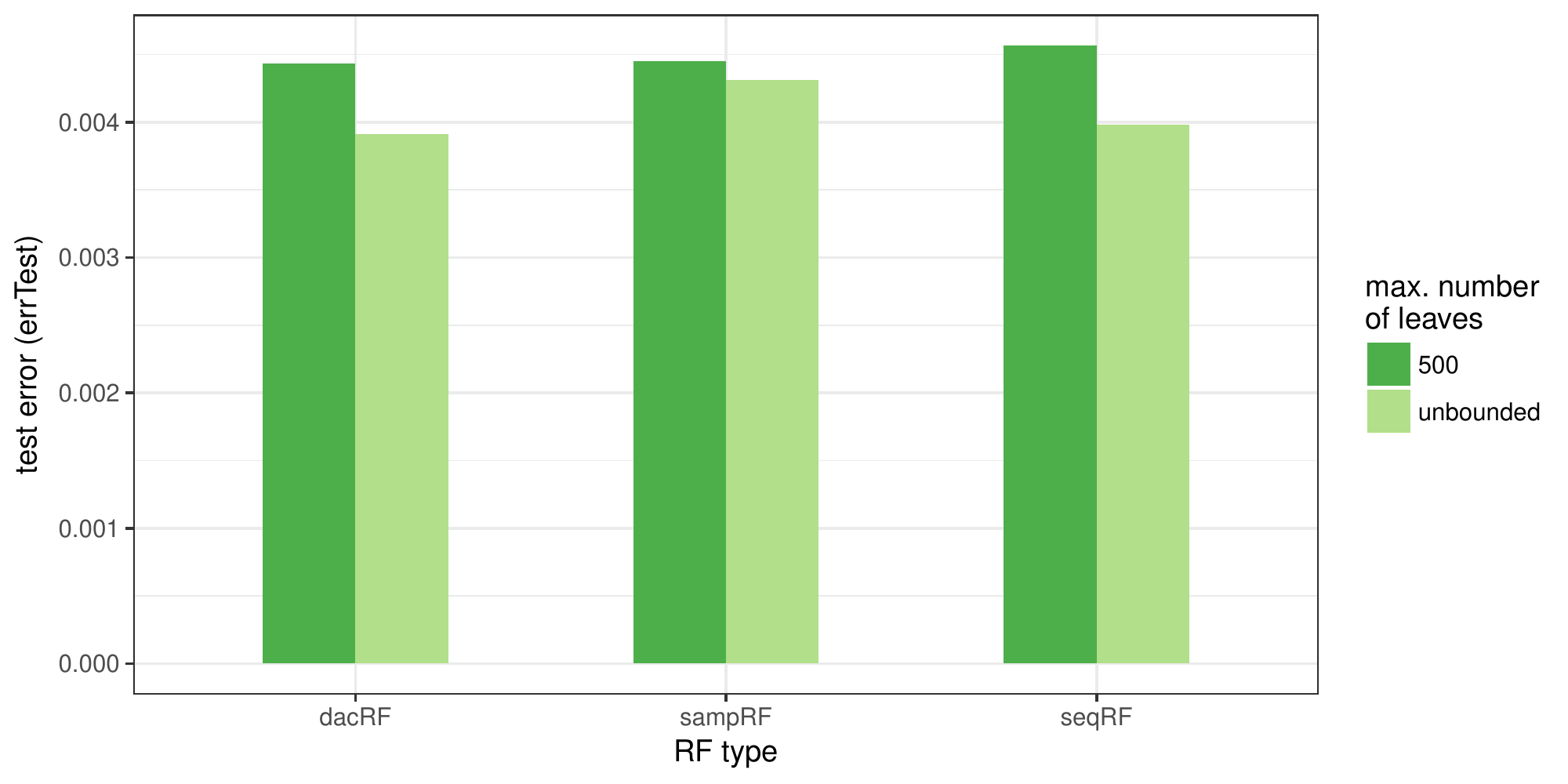}
	\caption{Prediction error (mesured by errTest) behavior for 3 RF methods when
	using maximal	trees or a maximum number of leaves of 500.}
	\label{fig::maxtrees}
\end{figure}

Finally, performance of 3 RF methods using maximal trees instead of
500 leaves trees were obtained. The results are illustrated in 
Figure~\ref{fig::maxtrees}.
Computational times are comparable to those shown in
Figures~\ref{fig::blb_mrrf_nbtrees} and \ref{fig::moonrf_samprf}, while the
misclassification rates are slightly better.
The remaining heterogeneity, when developing trees with 500 leaves, does
not affect much the performance in that case. Hence, while pruning all trees
would lead to a prohibitive computational time, a constraint on tree size may 
well be
adapted to the Big Data case. This point needs a more in-depth analysis and is
left for further research.

\subsection{Online random forest}
\label{simu-online-rf}

This section is dedicated to simulations with online RF. The simulations were 
performed with the method described in \cite{denil_etal_ICML2013} which is 
available at \url{https://github.com/david-matheson/rftk} ({\bf onRF}). The 
method is implemented in python. Thus computational time cannot be directly 
compared to the computational described in the two previous sections (because 
of the programming language side effect). Similarly, the input hyperparameters 
of \texttt{randomForest} function in the \RR{} package \pkg{randomForest} are 
not exactly the same than the ones proposed in {\bf onRF}: for instance, in the 
\RR{} package, the complexity of each tree is controlled by setting the maximum 
number of leaves in a tree whereas in {\bf onRF}, it is controlled by setting 
the maximum depth of the trees. Additionally, the two tools are very 
differently documented: every function and option in the \RR{} package are 
described in details in the documentation whereas RFTK is not provided with a 
documentation. However, the meaning of the different options and outputs of the 
library can be guessed from their names in most cases. 

When relevant, we discuss the comparison between the standard approaches tested 
in the two previous sections and the online RF tested in the current version but 
the reader must be aware that some of the differences might come directly from 
the method itself (standard or online), whereas others come from the 
implementation and programming languages and that it is impossible to 
distinguish between the two in most cases. 

The simulations in this section were performed on the datasets described in 
Section~\ref{exp-framework}. The training dataset (randomly ordered) took 
approximately 9 minutes to be loaded with the function \texttt{loadtxt} of the 
python library \pkg{numpy}, which is about 9 times larger than the time needed 
by the \RR{} package \pkg{readr} to perform the same task. In the sequel, 
results about this dataset will be referred as \textbf{standard}. Moreover, 
simulations were also performed to study the effect of sampling (subsamples 
drawn at random with a sampling fraction in $\{0.01,0.1,1,10\}$\%) or of biased 
order of arrival of the observations (with the datasets {\bf unbalanced}, with 
$p=0.01$, and {\bf x-biases} with 15 parts). For {\bf x-biases} the number of 
parts was chosen differently than in the Section~\ref{simulated} (for {\bf 
dacRF}) because only 2 parts would have led to a quite extreme situation for 
{\bf onRF}, in which all data coming from submodel 1 are presented first, 
before 
all data coming from submodel 2 are presented. We have thus chosen a more 
moderate situations in which data from the two submodels are presented by 
blocks, alternating submodel 1 and submodel 2 blocks. Note that both simulation 
settings are similar, since {\bf dacRF} processes the different (biased in $X$) 
blocks in parallel.

The forests were trained with a number of trees equal to 50 or 100 (for 
approximately 500 trees, the RAM capacity of the server was overloaded) and with 
a control of the complexity of the trees by their maximum depth which was varied 
in $\{5, 10, 15, 50\}$. RFTK does not provide the online approximation of OOB 
error so the accuracy was assessed by the computation of the prediction error on 
the same test dataset used in the previous two sections.

\begin{figure}[H]
	\centering
	\includegraphics[width=\linewidth]{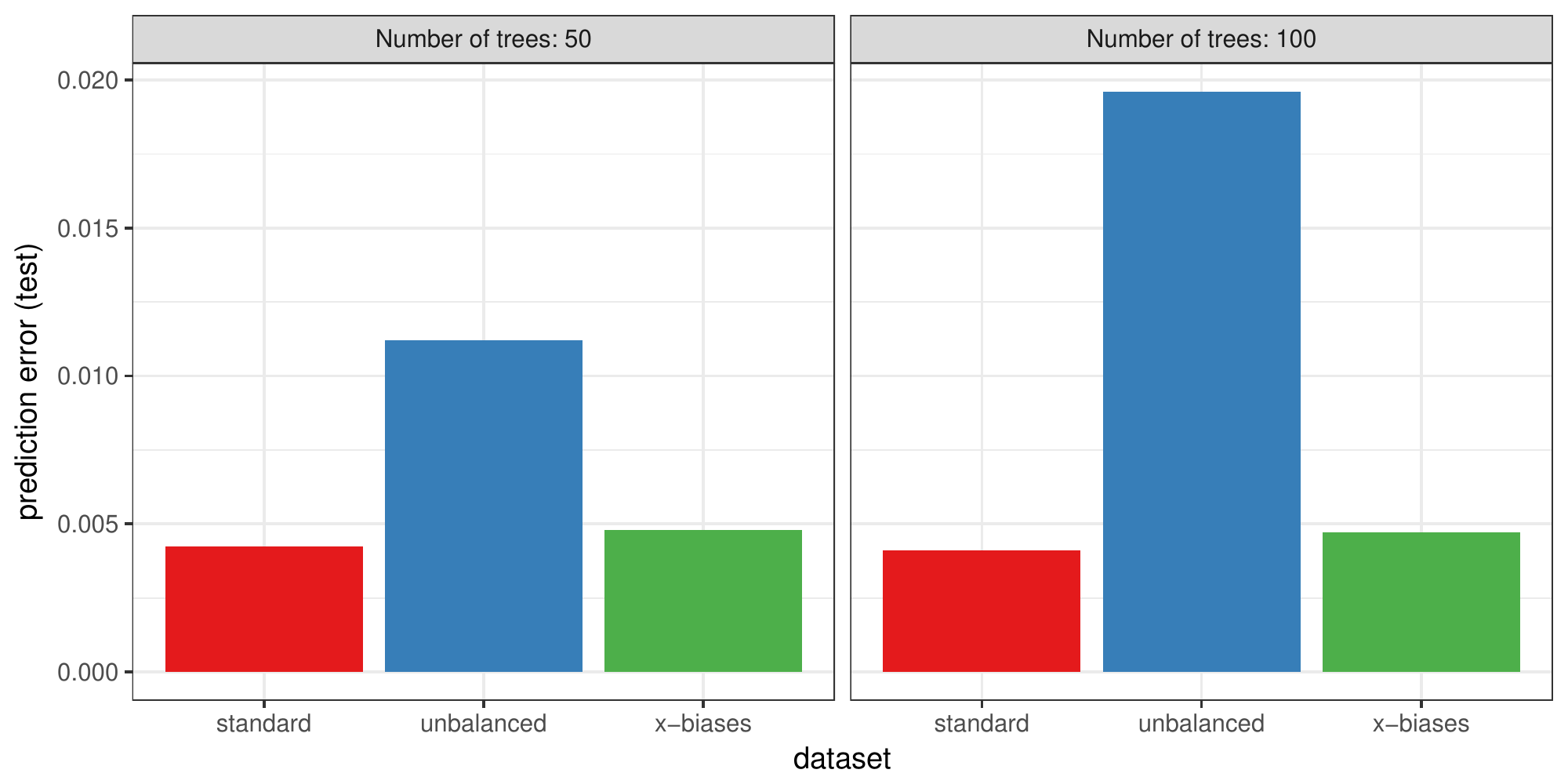}
	\caption{{\bf onRF}: Prediction error for the test dataset.}
	\label{fig::rftk_barplot-Acc}
\end{figure}

Figure~\ref{fig::rftk_barplot-Acc} displays the misclassification rate of {\bf 
onRF} on the test dataset versus the type of bias in the order of arrival of 
data (no bias, {\bf unbalanced} or {\bf x-biases}) and versus the number of 
trees in the forest. The results are provided for forests in which the maximum 
depth of the trees was limited to 15 (which almost always correspond to fully 
developed trees).
The result shows that, contrary to the {\bf dacRF} case, \textbf{x-biases} 
almost do not affect the accuracy of the results, even if the classifier always 
has a better accuracy when data are presented in random order. On the contrary, 
\textbf{unbalanced} has a strong negative impact on accuracy of the classifier. 
Finally, for the best case scenario ({\bf standard}), the accuracy of {\bf 
onRF} is not much affected by the number of trees in the forest but the 
accuracy tends to get even worse when increasing the number of trees in the 
worst case scenario ({\bf unbalanced}). In comparison with the strategies 
described in Section~\ref{simulated}, {\bf onRF} has comparable test error 
rates (between $(4-4.3)\times 10^{-3}$) for forests with 100 trees).

\begin{figure}[ht]
	\centering
	\includegraphics[width=\linewidth]{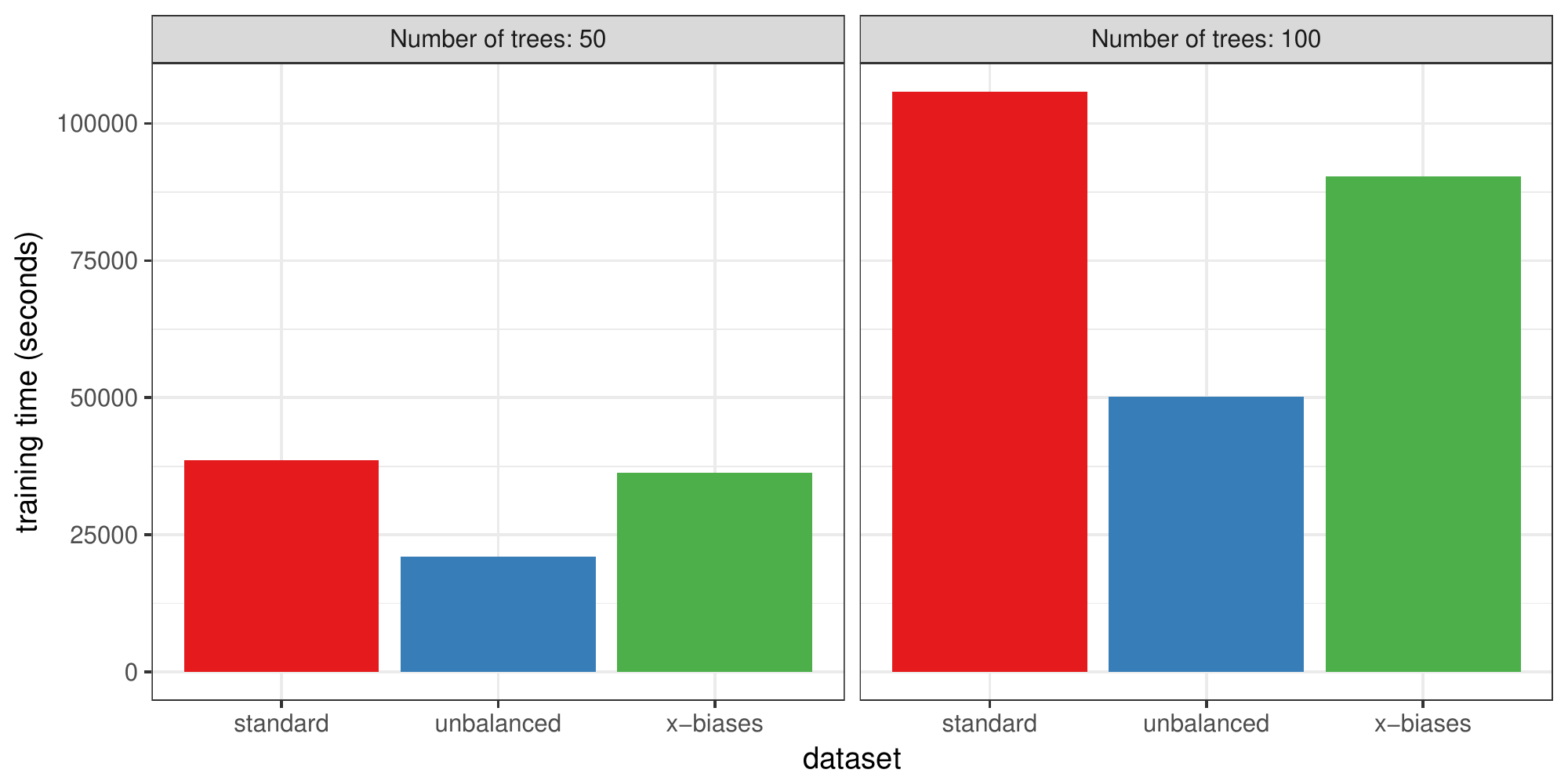}
	\caption{Training time (seconds) of {\bf onRF}.}
	\label{fig::rftk_barplot-CT}
\end{figure}

Additionally, Figure~\ref{fig::rftk_barplot-CT} displays the evolution of the 
computational time  versus the type of bias in the order of arrival of 
data and the number of trees in the forest. The results are provided for forests 
in which the maximum depth of the trees was limited to 15.
As expected, computational time increases with the number of trees in the 
forest (and the increase is larger than the increase in the number of trees). 
Surprisingly, the computational time of the worse case scenario ({\bf 
unbalanced} bias) is the smallest. A possible explanation is the fact that 
trees are presented successively a large number of observations with the same 
value of the target variable ($Y$): the terminal nodes are thus maybe more 
easily pure during the training process in this scenario.

Computational times are hard to compare with 
the ones obtained in Section~\ref{simulated}. However, computational times 
 are of order 30 minutes at most for {\bf 
dacRF}, and 1-2 minutes for {\bf blbRF} and {\bf moonRF}, whereas {\bf onRF} 
takes approximately 10 hours for 50 trees and 30 hours for 100 trees, which is 
even larger than training the forest sequentially with \pkg{randomForest} (7 
hours).

\begin{figure}[ht]
	\centering
	\includegraphics[width=\linewidth]{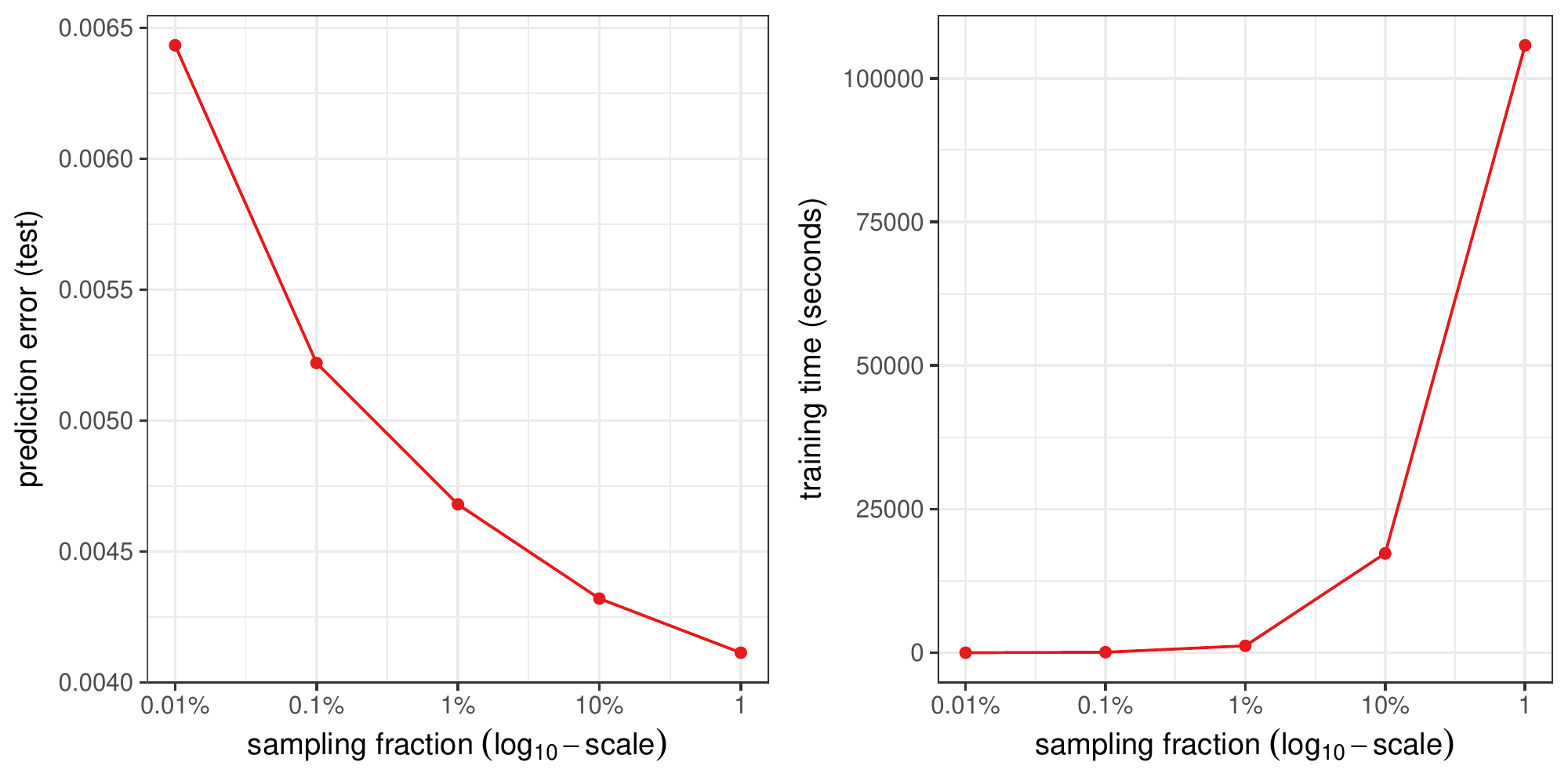}
	\caption{Prediction error (left) and training time (right) versus sampling 
fraction for {\bf onRF}. $x$-axis is $\log_{10}$-scaled.}
	\label{fig::rftk_sampling}
\end{figure}

Figure~\ref{fig::rftk_sampling} displays the evolution of the misclassification 
rate and of the computational time versus the sampling fraction when a random
subsample of the dataset is used for the training (the number of trees in the 
forest is equal to 100 and the maximum depth set to 15). The computational time 
needed to train the model is more than linear but the prediction accuracy also 
decreases in a more than linear way with the sampling fraction. The loss in 
accuracy is slightly worse than what was obtained in Section~\ref{simulated} for 
{\bf sampRF}, showing than {\bf onRF} might need a 

\begin{figure}[H]
	\centering
	\includegraphics[width=\linewidth]{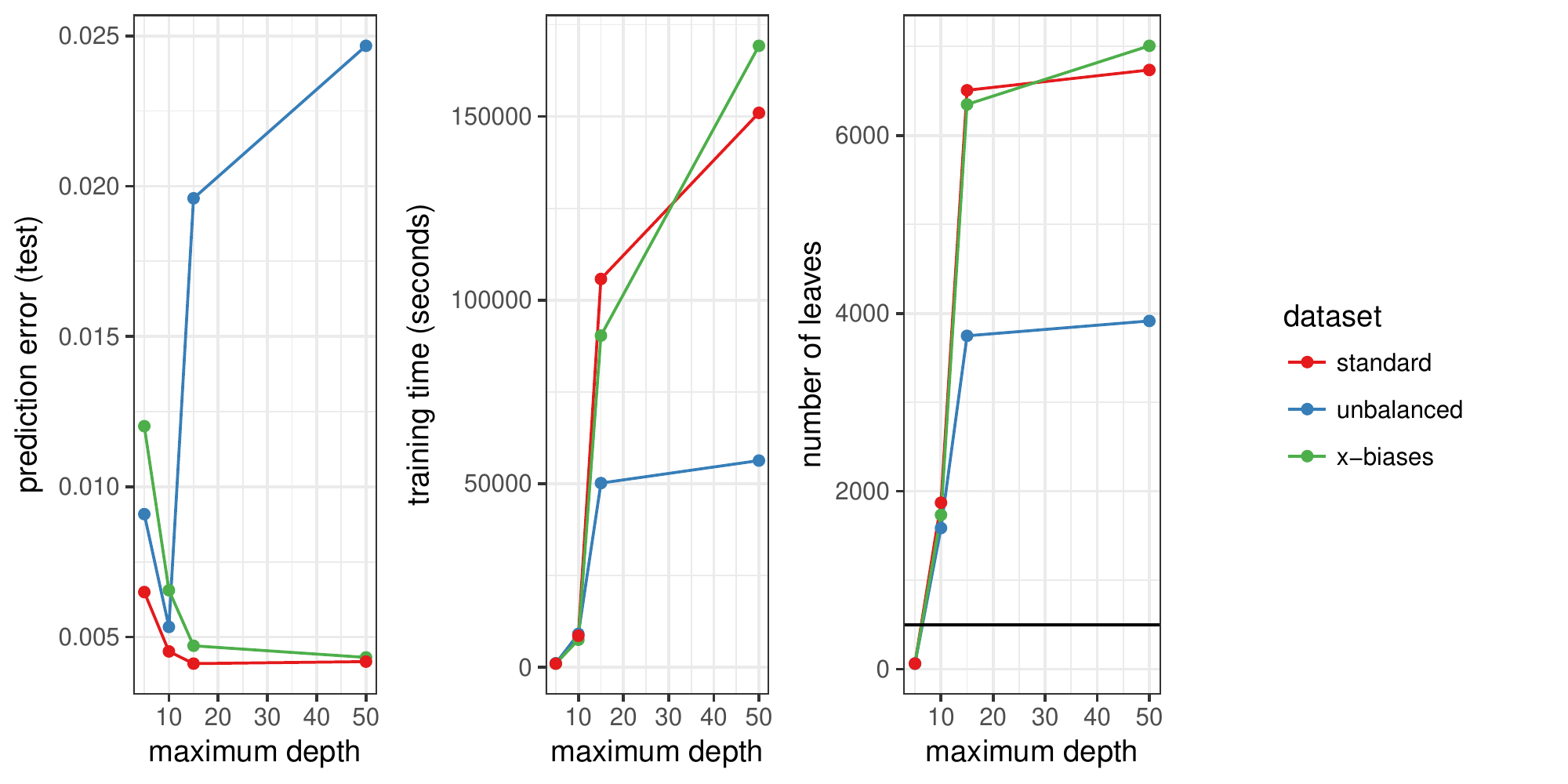}
	\caption{Top: Average depth of trees in the forest. Bottom: Average number of 
leaves of trees in the forest. The black horizontal line corresponds to the 
maximum number of leaves used in experiments of Sections~\ref{simulated} and 
\ref{simulated-biases}.}
	\label{fig::rftk_depth}
\end{figure}

Finally, Figure~\ref{fig::rftk_depth} displays the evolution of the test 
misclassification rate, of the computational time and of the average number of 
leaves in the trees versus the value of the maximum depth for forests with 100 
trees. As expected, the computational time is in direct relation with the 
complexity of the forest (number of trees and maximum depth) but tends to 
remain almost stable for trees with maximum depth larger than 15. The same 
behavior is observed for the misclassification rate in {\bf standard} and {\bf 
x-biases} which reach their minimum for forests with a maximum depth set to 
15. Finally, the number of leaves for \textbf{unbalanced} is much smaller, 
which also explains why the computational time needed to train the forest in 
this case is smaller. For this type of bias, the misclassification rates 
increases with the maximum depth for forest with maximum depths larger than 10: 
as for the number of trees, the complexity of the model seem to have a negative 
impact on this kind of bias.

\subsection{Airline dataset}
\label{airlines}

In the present section, similar experiments are performed with a real world 
dataset related to flight delays. The data were first processed in 
\cite{kane_etal_JSS2013} to illustrate the use of the \RR{} packages for Big 
Data computing \pkg{bigmemory} and \pkg{foreach} \cite{RA_weston_R2014}. In 
\cite{kane_etal_JSS2013}, the data were mainly used for description purpose 
({\it e.g.}, quantile calculation), whereas we will be using it for prediction. 
More precisely, five variables based on the original variables included in the 
data set were used to predict if the flight was likely to arrive on time or 
with 
a delay larger than 15 minutes (flights with a delay smaller than 15 minutes 
were considered on time). The predictors were: the moment of the flight (two 
levels: night/daytime), the moment of the week (two levels: weekday/week-end), 
the departure time (in minutes, numeric) and distance (numeric). The dataset 
used
to make the simulations contained 120,748,239 observations (observations with 
missing values were filtered out) and had a size equal to 3.2 GB (compared to 
the 12.3 GB of the original data with approximately the same number of 
observations). Loading the dataset and processing it to compute and extract the 
predictors and the target variables took approximately 30 minutes. Another 
feature of the dataset is that it is unbalanced: most of the flight are on time 
(only 19.3\% of the flights are late).

The same method than the one described in Section~\ref{simulated} were compared:
\begin{itemize}
	\item a standard RF, {\bf seqRF}, was computed sequentially.
	It contained 100 trees. The RF took 16 hours to be obtained and its OOB error
	was equal to 18.32\%;
	\item {\bf sampRF} was trained with a subsample of the total data
	(1\% of all the observations were sampled at random without replacement).
	These RF were trained in parallel with 15 cores, each core building 7 trees 
	from bootstrap samples coming from the common subsample (the final RF hence
	contained 105 trees);
	\item a {\bf blbRF} was also trained using $K=15$ subsamples,
	each containing about 454,272 observations (about 0.4\% of the size of the 
	total data set). 15 sub-forests were trained in parallel with 7 trees each
	(the final forest hence contained 105 trees);
	\item Finally {\bf dacRF} was also obtained with $K=15$ chunks and $q=7$ 
	trees in each sub-forest grown in the different (the final RF contained from
	to 1000 trees).
\end{itemize}

The number of trees, $q$, built in each chunk for {\bf dacRF} is smaller than
what seemed a good choice in Section~\ref{simulated}, but for this example,
increasing the number of trees did not lead to better accuracy (even if it
increased a lot the computational time).

\begin{figure}
\centering
	\includegraphics[width=\linewidth]{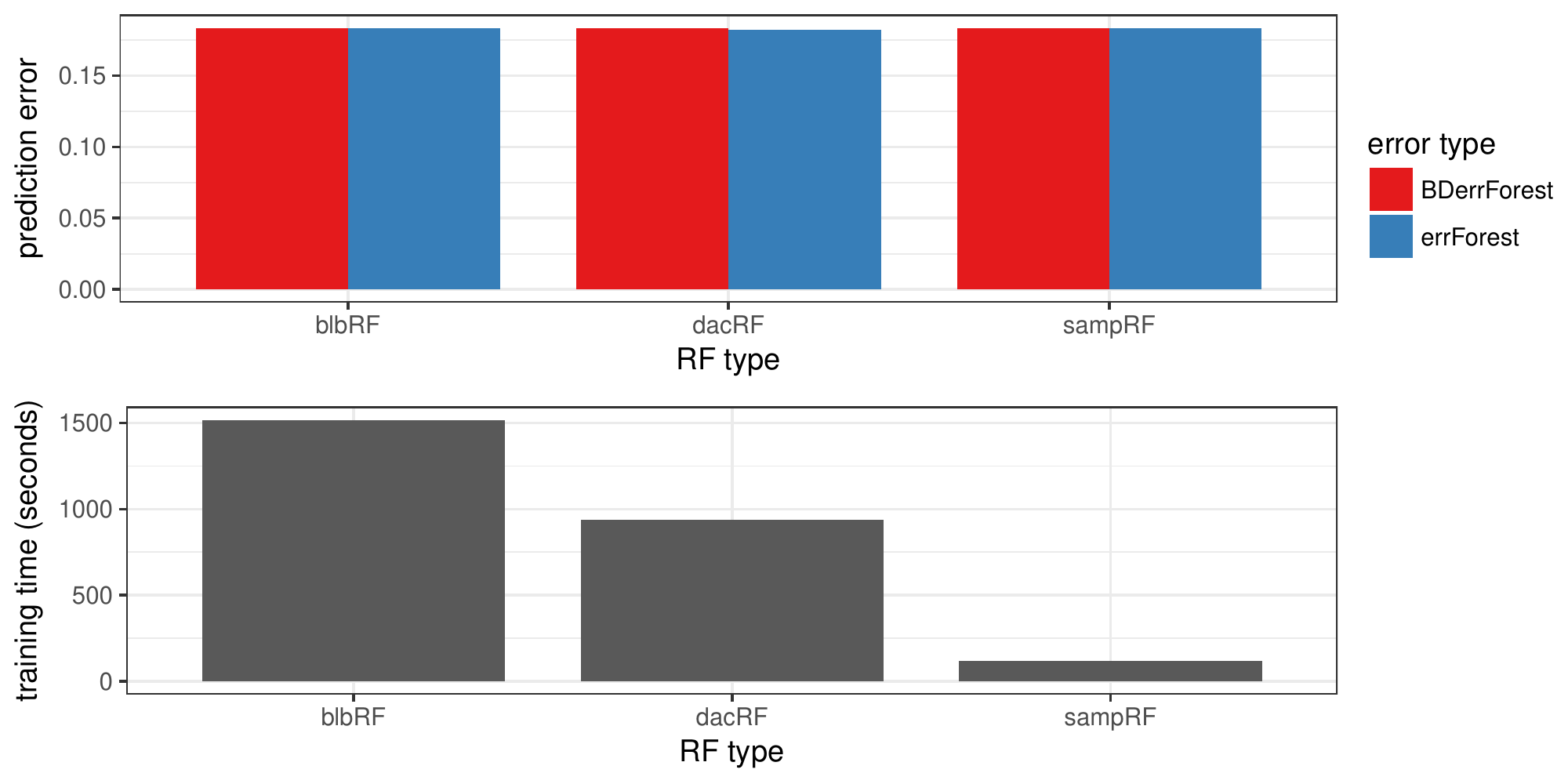}
	\caption{Performance (computational time and misclassification rates) 
obtained by three different RF methods for Big Data on Airline data.}
	\label{fig::airline}
\end{figure}

In all methods, the maximum number of terminal leafs in the trees was set to 
500 and all RF were trained in parallel on 15 cores, except for the sequential 
approach. Results are given in Figure~\ref{fig::airline} in which the 
notations are the same as in Section~\ref{simulated}.
The results show that there is almost no difference in terms of performance 
accuracy between using all data and using only a small proportion (about 
0.01\%) 
of them. In terms of compromise between computational time and accuracy, using 
a 
small subsample is clearly the best strategy, provided that the user is able to 
obtain a representative subsample at a low computational cost. Also, contrary 
to 
what happened in the example described in Section~\ref{simulated-biases}, 
BDerrForest is always a good approximation of errForest.
An explanation of this result might be that for Airline dataset, prediction
accuracy is quite poor and this might be due to explanatory variables that
are not
informative enough. Hence differences between BDerrForest and errForest may
be hidden by the fact that the two estimations of the prediction error 
are quite
high.

In addition, the impact of the representativity, with respect to the target 
variable, of the samples on which the RF were trained was assessed: instead of 
using a representative (hence unbalanced) sample from the total dataset, a 
balanced subsample (for 50\% of delayed flights and 50\% of on time flights) 
was obtained and used as the input data to train the random forest. Its size 
was equal to 10\% of the total dataset size. This approach obtained an 
errForest equal to 33.34\% (and BDerrForest was equal to 39.15\%), which is 
strongly deteriorated compared to the previous misclassification rates. In this 
example, the representativity of the observations contained in the subsample 
strongly impacts the estimated model. The model with balanced data has a better 
ability to detect late flights and favors the sensitivity over the specificity.

\section{Conclusion and discussion}
\label{conclusion}

This final section provide a short conclusion and opens two perspectives.
The first one proposes to consider re-weighting random forests as an alternative for 
tackling the lack of representativeness for BD-RF and the second one focuses on
alternative online RF schemes as well RF for data streams. 

\subsection{Conclusions}

This paper aims at extending standard Random Forests in order to process Big 
Data. Indeed RF is an interesting example among the widely used statistical methods 
in machine learning since it already offers several ways to deal with massive 
data in offline or online contexts. Focusing on classification problems, we 
reviewed some of the available proposals about RF in parallel environments and 
online RF. We formulated various remarks for RF in the Big Data context, 
including approximations of out-of-bag type errors. We experimented on two 
massive datasets (15 and 120 millions of observations), a simulated one and real 
world data, five variants involving subsampling, adaptations of bootstrap to Big 
Data, a divide-and-conquer approach and online updates. 

Among the variants of RF that we tested, the fastest were {\bf sampRF} with a 
small sampling fraction and {\bf blbRF}. On the contrary, {\bf onRF} was not 
found computationally efficient, even compared to the standard method {\bf 
seqRF}, in which all data are processed as a whole and trees are built 
sequentially. On a performance point of view, all methods provide satisfactory 
results but parameters (size of the subsamples, number of chunks...) must be 
designed with care so as to obtain a low prediction 
error. However, since the estimation of OOB error that can be simply designed 
from the different variants was found a bad estimate of the prediction error in 
many cases, it is also advised to rather calculate an error on an independent 
smaller test subsample. When the amount of data is that big, computing such a 
test error is easy and can be performed at low computational cost. 

Finally, one of the most crucial point stressed in the simulations is that the 
lack of representativeness of subsamples can result in drastic deterioration of 
the performances of Big Data variants of RF, especially of {\bf dacRF}. 
However, designing a subsample representative enough of the whole dataset can be 
an issue {\it per se} in the Big Data context, but this problem is out of the 
scope of the present article. 

\subsection{Re-weighting schemes}

As an alternative, some re-weighting schemes could be used to address the issue of the 
lack of representativeness for BD-RF. Let us sketch some possibilities.

Following a notation from Breiman \cite{breiman_ML2001}, RF lead to better results when there 
is a higher diversity among the trees of the forest. So recently, some 
extensions of RF have been defined for improving an initial RF. In
\cite{fawagreh_etal_p2015}, Fawagreh {\it et al.} use an unsupervised learning 
technique (Local Outlier Factor, LOF) to identify diverse trees in the RF and 
then, they perform ensemble pruning by selecting trees with the highest LOF 
scores to produce an extension of RF termed LOFB-DRF, much smaller in size than 
RF and performing better. This scheme can be extended by using other diversity 
measures, see \cite{tang_etal_ML2006} presenting a theoretical analysis on six 
existing diversity measures.

Another possible variant would be to consider the whole forest as an ensemble 
of forests and to adapt the majority vote scheme with weights that address, 
{\it e.g.}, the issue of the sampling bias. Recently in 
\cite{winham_etal_SADM2013}, Winham {\it et al.} propose to introduce a 
weighted RF approach to improve predictive performance: the weighting scheme is 
based on the individual performance of the trees and could be adapted to the 
{\bf dacRF} framework.

Along the same ideas it would be, at least for an exploratory stage, possible to adapt a simple 
idea coming from the variants of AdaBoost 
\cite{freund_schapire_JCSS1997} for classification boosting algorithms. Recall 
that the basic idea of boosting is, as for the RF case, to generate many different base predictors obtained by 
perturbing the training set and to combine them. Each predictor is designed 
sequentially highlighting the observations poorly predicted. This is a crucial
difference 
with RF scheme for which the different training samples are obtained by independent 
bootstraps. But the aggregation part of the algorithm is interesting here: instead of taking 
the majority vote of the trees predictions as in the RF context, a weighted combination of trees is considered.
The unnormalized weight of the tree $t$ is simply $\alpha_t=1/2\ln(\epsilon_t/(1-\epsilon_t))$ 
where $\epsilon_t$ is the misclassification 
error computed on the whole training sample $L$. This could be adapted by considering weighted forests 
using weights of such form, evaluated on a same (small) subset of observations supposed to be representative 
of the whole dataset. 

\subsection{Online data and Data Streams}

The discussion sketched about online RF can be extended. Indeed the use of ERT variant of RF
instead of Breiman's RF allows to reduce the computational cost. It would be of 
interest to use this RF variant in {\bf dacRF}, or even more randomized ones 
(like \cite{cutler_zhao_CSS2001} PERT, Perfect Random Tree Ensembles, or 
\cite{biau_etal_JMLR2008,arlot_genuer_p2014} PRF, Purely Random Forests).
The idea of those latter variants is to not choose the variable involved in a 
split and the associated threshold from the data but to randomly choose them 
according to different schemes. Finally, {\bf onRF}
could be a way to use only a portion 
of the data set until the forest is accurate enough. Moreover, one valuable
characteristic of {\bf onRF} is that it could address both the issue of Volume 
and Velocity.

In the framework of online RF, only sequential inputs are considered. But more 
widely in the Big Data context, data streams are of interest. They allows to
consider not only sequential inputs, but  also entail unbounded data that 
should be processed in limited (given their unboundedness) memory and in 
an online fashion to obtain real-time answers to application queries 
(for an accurate and formal one, see \cite{garofalakis_etal_DSMPHSDS2016}).
Moreover, data streams can be processed in observation- or time-based windows 
or even batches which collect a number of recent observations (see for instance 
\cite{giannella_etal_DMNGCFD2004}). It could be interesting to
fully adapt online RF to the data stream context (see for example 
\cite{abdulsalam_etal_IEEETKDE2011} 
and \cite{abdulsalam_etal_DEXA2008}) and  obtain similar theoretical
results.


\section*{Additional Files}
  \subsection*{Additional file 1 --- \RR{} and python scripts used for the 
simulation}

	\RR{} scripts used in the simulation sections are available at 
\url{https://github.com/tuxette/bigdatarf}.

\end{document}